\documentclass{article}

\usepackage[preprint]{corl_2026} 
\usepackage{amsmath}
\usepackage{amssymb}
\usepackage{algorithm}
\usepackage[noend]{algpseudocode}
\usepackage{graphicx}
\usepackage{wrapfig}
\usepackage{float}
\usepackage{pdflscape}
\usepackage{caption}
\usepackage{subcaption}
\usepackage{titlesec}
\usepackage{xspace}

\titlespacing*{\section}{0pt}{0.8\baselineskip}{0.3\baselineskip}
\titlespacing*{\subsection}{0pt}{0.6\baselineskip}{0.2\baselineskip}
\titlespacing*{\subsubsection}{0pt}{0.5\baselineskip}{0.2\baselineskip}
\newcommand{\MethodName}{\emph{Freeform Preference Learning}\xspace}
\newcommand{\Acronym}{\emph{FPL}\xspace}

\usepackage[most]{tcolorbox}
\tcbset{
  promptbox/.style={
    enhanced,
    colback=gray!4,
    colframe=black!55,
    boxrule=0.6pt,
    arc=6pt,          
    left=10pt,right=10pt,top=8pt,bottom=8pt,
    fonttitle=\bfseries,
    coltitle=black,
    attach boxed title to top left={xshift=8pt,yshift=-2mm},
    boxed title style={
      colback=white,
      colframe=black!55,
      arc=4pt,
      boxrule=0.6pt,
      left=6pt,right=6pt,top=2pt,bottom=2pt
    }
  }
}

\newtcolorbox{PromptBox}[1]{promptbox,title={#1}}

\title{Freeform Preference Learning for Robotic Manipulation}
\author{
Marcel Torne\thanks{Equal contribution} \quad
Anubha Mahajan\footnotemark[1] \quad
Abhijnya Bhat\footnotemark[1] \quad
Chelsea Finn \\
\\
Stanford University
}

\begin{document}
\maketitle

\begin{abstract}
    Reward design remains a central bottleneck for autonomous robot policy improvement, especially in long-horizon manipulation tasks where sparse success labels provide too little signal and binary preferences collapse many competing notions of quality into one ambiguous signal. We introduce \MethodName{} (\Acronym), a method for learning robot policies from freeform human preferences. Rather than asking annotators which of two trajectories is better overall, \Acronym{} lets them define natural-language preference axes, such as speed, safety, quality of placement, or carefulness, and provide pairwise preferences along each axis. These annotations are used to learn a language-conditioned reward model that maps a trajectory and preference label to an axis-specific reward. We use this model to train a reward-conditioned policy that optimizes across the multiple human-specified dimensions. Across four real-world and two simulated long-horizon manipulation tasks, \Acronym{} improves over sparse-reward and binary-preference methods by 38 percentage points. Beyond improved performance, \Acronym{} learns dense progress signals without explicit subtask segmentation, shows compositionality of behavior not present in the data, and allows users to steer the policy towards different behaviors at test time without retraining. Blog post with videos available at \nolinkurl{freeform-pl.github.io/fpl.website/}
\end{abstract}



\section{Introduction}

Rewards are a critical component for autonomous robot improvement. An ideal reward function should provide dense, unambiguous feedback and should capture all aspects of desirable behavior. For example, supervision for the simple task of setting a table should incorporate the configuration of the cutlery, the degree of care taken to not break fragile plates, the comfort of nearby people (e.g. to avoid motions that point a knife towards a person), and the speed of execution, among other aspects. Accurately capturing all of these axes presents a major challenge, both when eliciting supervision from people and when representing all of these factors in a reward function and downstream behavior. Moreover, a reward function that captures these axes but is too sparse, or dense but inaccurate, can lead to unwanted downstream behaviors when optimized against. In this paper, we study how to leverage human supervision to learn reward functions and ultimately robot behavior that captures all dimensions of a person’s intent.

Prior works have studied a variety of rewards and reward learning approaches. Perhaps the simplest option is to provide or learn from binary success labels \cite{luo2024serl,singh2019end,kalashnikov2018scalable}, which should in principle make it easy for people to determine if all criteria are met. However, this reward signal places significant burden on the reinforcement learning algorithm, making it hard to scale to more challenging tasks and to incorporate real-world constraints on behavior beyond basic task completion. Other works learn shaped scalar rewards \cite{liang2026robometer,lee2026roboreward,chen2026topreward} but still focus on task progress metrics, ignoring important criteria on how a task was performed. Finally, preference learning \cite{christianoHumanPref,zhang2024grape,Sadigh2017ActivePL} is a promising paradigm for learning denser reward signals while reducing the burden on human supervisors, but requires them to collapse multiple axes of judgment into a single ``overall" binary preference label. In long-horizon tasks this can make preferences difficult to provide and the resulting supervision ambiguous.

\begin{figure}[!t]
    \centering
      \includegraphics[width=\linewidth]{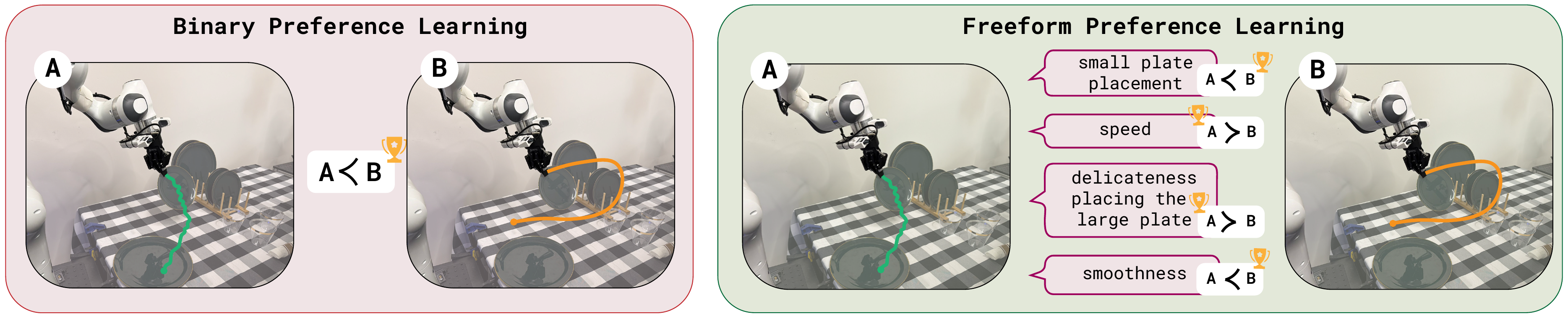}
    \vspace{-1.5em}
    \caption{\footnotesize Comparison between binary preference learning (\emph{left}) and \MethodName~(\emph{right}). Binary preference learning asks annotators to compare two trajectories using a single ``overall preference''. \MethodName~ instead collects more detailed feedback by letting the annotator specify the axes in natural-language and provide per-axis preferences.}
    \label{fig:main}
    \vspace{-1.5em}
\end{figure}

To preserve these axes of judgment, our key insight is to allow annotators to provide preference labels on any task-relevant axes of their choosing. This multidimensional supervision can then be used to learn a policy optimized for the combination of these dimensions. We instantiate this idea by asking annotators to specify relevant judgment dimensions in natural language and to provide a binary preference for each axis. The axes can be defined up front or during the annotation process. We then learn a multi-axis reward function that produces a scalar reward score when conditioned on a natural language description of the axis. Finally, we train a promptable policy to optimize the combination of axes described during reward training. Notably, this framework simultaneously improves both the ease of providing unambiguous supervision and the density of supervision for downstream policy optimization.

The main contribution of our paper is \MethodName (\Acronym), a method for eliciting and learning from freeform human preferences. \Acronym learns a language-conditioned reward function from preferences, capturing a variety of task-relevant attributes including quality of result, speed, smoothness, damage, and hygiene. This reward model provides dense supervision for training a multi-axis reward-conditioned policy. We evaluate \Acronym on four real-world tasks—putting a cube in a target bowl, folding shorts, plating a toast, and setting up a table—as well as two simulated tasks. Across settings, policies trained with \Acronym significantly outperform those trained with sparse rewards and binary preference learning methods. We further find that preserving the multi-dimensionality of feedback enables compositional generalization and test-time steerability of the resulting policies as well as qualitatively denser rewards on long-horizon tasks without requiring subtask segmentation.

\section{Preliminaries}
In this section, we will give an overview of learning from binary human preferences and, with particular emphasis on the foundational work of \citet{christianoHumanPref}.

\textbf{Reward learning from binary human feedback}:
Learning from binary human feedback offers a way to provide dense reward signal by learning a reward from human preferences. This avoids the challenge of designing a dense reward based on images. When learning from human feedback, two states or segments $(s_i, s_j)$ from two different trajectories are presented to the annotators. The annotator is then asked to indicate their preference $y \in \{0, 1\}$ between the two, where we define \(y=1\) if the annotator prefers \(s_i\) over \(s_j\), and \(y=0\) otherwise. Multiple such pairs are compared and form the dataset of pairwise preferences, $\mathcal{P}$. The Bradley-Terry model assumes that the probability of preferring $s_i$ over $s_j$, where $\sigma(\cdot)$ is the logistic function, is:
\begin{equation}
\label{eq:bt-states}
P(s_i \succ s_j) = \frac{\exp\!\left(r_\phi(s_i)\right)}{\exp\!\left(r_\phi(s_i)\right) + \exp\!\left(r_\phi(s_j)\right)} = \sigma\!\left(r_\phi(s_i) - r_\phi(s_j)\right),
\end{equation}
 The reward model $r_\phi$ is then trained by minimizing the negative log-likelihood of the preferences:
\begin{equation}
\label{eq:bt-states-loss}
\mathcal{L}_{\text{BT}}(\phi) = -\mathbb{E}_{(s_i,\, s_j,\, y) \sim \mathcal{P}} \left[ \log \sigma\!\bigl((2y-1)\left[~r_\phi(s_i) - r_\phi(s_j)\right]\bigr) \right],
\end{equation}
In practice, providing binary preferences is ambiguous and noisy. In Section \ref{sec:fplmethod} we propose a new method to learn from freeform preferences instead.

\textbf{Policy extraction from the learned reward}:
After learning the reward model, we must extract the policy that maximizes the reward. \citet{christianoHumanPref} proposed using Proximal Policy Optimization \cite{schulmanppo}, an online on-policy RL algorithm. Using this class of algorithms has proven to be effective for robotics in simulation, as well as for large language models \cite{ouyang2022training} where obtaining rollout data is not costly. In practice, for real-world robot learning, PPO is too sample inefficient. 
In Section \ref{sec:fplmethod}, we propose using another class of policy extraction algorithms, off-policy and batch online, that make real-world RL tractable.

\section{Learning from Freeform Preferences}
\label{sec:fplmethod}

The key idea behind our method, \MethodName (\Acronym), is to learn from natural language and open-ended feedback instead of traditional binary preferences. Rather than asking annotators for one overall preference between two trajectories, we ask them to describe the axes along which to compare them, such as speed, safety, smoothness, or subtask completion. This yields feedback that is more granular and less ambiguous. We use these freeform preferences to learn a language-conditioned, multi-dimensional reward function that scores trajectories along each specified axis. We then train a policy conditioned on multiple preference dimensions to optimize the behavior with respect to each axis. We describe each component of our proposed algorithm below.

\subsection{Learning a Reward Function from Freeform Human Feedback}
Traditionally, human preferences are collected as a binary signal over an ``overall quality" metric, i.e., the annotators are asked ``which one of the two trajectories do you prefer?" However, this signal is difficult to provide because the answer often depends on the axis of comparison. For example, if one trajectory is faster while the other is safer, it is unclear which should be preferred overall. This ambiguity is further amplified when preferences are collected over trajectory segments, since the two segments may correspond to different stages of the task and therefore be difficult to directly compare. 
We instead collect freeform human preferences by showing two full trajectories and asking them to evaluate them along multiple axes, either predefined or specified by the annotator in natural language. 
Rather than asking a single fixed question about  ``overall quality", we collect preferences over a variety of axes such as ``formality of setup", ``speed", ``safety", and so on. 
Given a freeform preference dataset $\mathcal{P}$ made up of natural language labels defining the axes $l_k$ and binary preferences $y_k$ per axis, we now describe how to learn a reward function $r_\phi$ from it.

\begin{figure}[t]
    \centering
    \includegraphics[width=\linewidth]{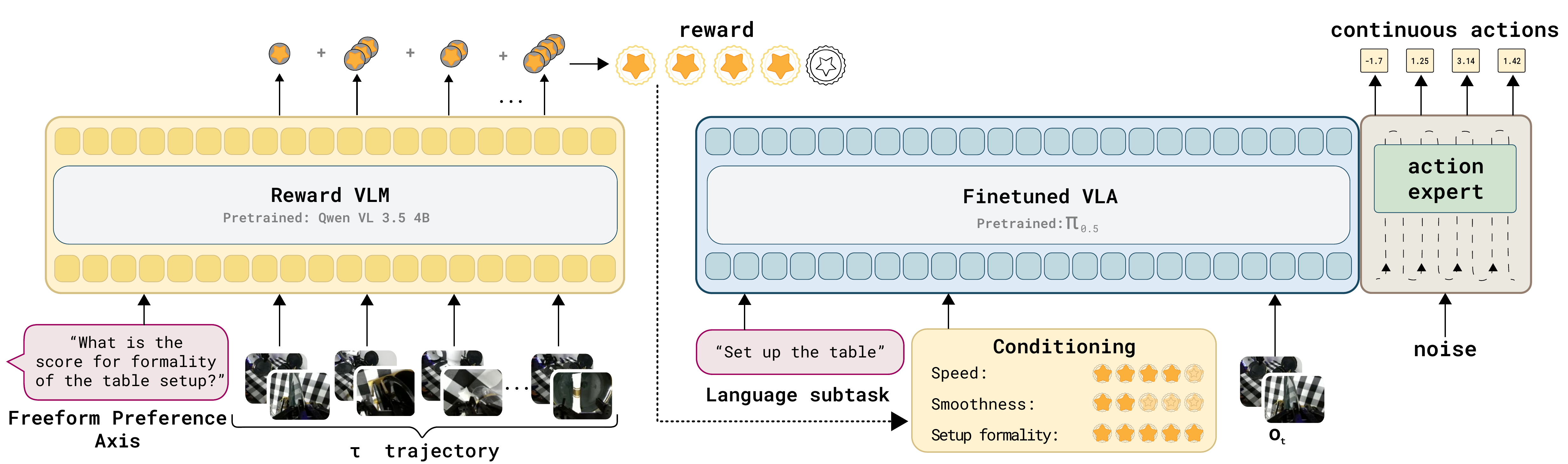}
    \vspace{-1.5em}
    \caption{\footnotesize \Acronym~ learns a multi-dimensional reward function to score the complete trajectory $\tau$ conditioned on the natural language preference axis. To leverage the multi-dimensionality of the reward function, \Acronym~ learns to reproduce behavior conditioned on the reward over multi-dimensional axes in text form. At test-time we can steer it towards high reward behaviors.}
    \label{fig:method}
    \vspace{-1.5em}
\end{figure}

A simple way to learn from multi-dimensional feedback would be to define a fixed set of $K$ preference axes and train a separate reward function for each one  \cite{wu2023fine}. However, this requires a predefined set of axes and limits generalization across semantically similar descriptions. For example, different annotators may refer to the same concept as ``speed", ``fast", or ``efficient". We instead keep preference axes in natural language and condition a single reward model directly on their text descriptions, leveraging the pretrained representations of vision-language models, see Fig. \ref{fig:method}.

The Bradley-Terry model in Eq. \ref{eq:bt-states-loss} is typically used to learn a uni-dimensional reward \cite{christianoHumanPref,pebble,torne2023breadcrumbs}. We extend this formulation to freeform preferences by conditioning the reward model on the natural-language axis for which each preference was provided. This yields an axis-conditioned reward function $r_\phi$ that scores a trajectory with respect to the specified axis, as defined in Eq \ref{eq:multi-axis-bt}. 

For a pair of trajectories $(\tau_i, \tau_j)$, an annotator provides per-axis preference labels and language labels $\{(l_k, y_k)\mid k=1,\ldots,K_{ij}\}$, where $y_k \in \{0,1\}$ indicates the preferred trajectory along axis $k$ and $K_{ij}$ is the total number of preference axes described for the trajectory pair (which may vary across pairs). The multi-dimensional reward model $r_\phi$ outputs a single scalar conditioned on the axis label and is trained to minimize the Bradley-Terry negative log-likelihood:
\vspace{-0.5em}
\begin{equation}
\label{eq:multi-axis-bt}
\mathcal{L}_{\text{\Acronym}-Reward}(\phi) = -\mathbb{E}_{(\tau_i,\, \tau_j,\, \{(l_k, y_k)\mid k=1,\ldots,K_{ij}\}) \sim \mathcal{P}} \left[ \sum_{k=1}^{K_{ij}} \log \sigma \biggl( (2y_k-1)\bigl(~ r_\phi(\tau_i | l_k) - ~ r_\phi(\tau_j | l_k) \bigr) \biggr) \right],
\end{equation}

Finally, unlike standard preference-based reward models that score individual observations or short segments~\cite{christianoHumanPref}, we condition the reward model on the observation history. This allows the model to capture temporal dependencies that may be necessary for evaluating trajectory-level preferences. Given a trajectory $\tau = {o_1,\ldots,o_T}$ and a natural-language preference axis $l_k$, we define the trajectory-level reward as a sum of per-prefix scores:
\vspace{-0.8em}
\begin{equation}
\label{eq:trajectory-reward}
r_\phi(\tau, l_k) = \sum_{i=1}^T g_\phi(o_{1:i}, l_k),
\end{equation}
where $g_\phi$ is a multimodal transformer that scores the segment $o_{1:i}$ conditioned on the axis $l_k$.

\subsection{Policy Extraction on a Multi-Dimensional Reward Function}

Now that we have described how to learn a language-conditioned reward function from freeform feedback on trajectory pairs, we discuss how best to use this supervision to learn a policy. One option would be to collapse the different dimensions into a scalar reward by a weighted sum \cite{hwang2024promptable} and optimize it with standard RL techniques. Although this is simple and would leverage the benefits of easier-to-provide feedback, policy optimization can also benefit significantly from more detailed supervision. Collapsing all the axes into a single scalar is more prone to reward hacking and exhibiting the problems of reward shaping \cite{chen2023visual}. Moreover, as we show in Section \ref{sec:experimentaleval}, preserving the decomposed rewards enables the learned policy to exhibit compositionality of behaviors not present in the original dataset, as well as at test-time steerability without retraining with a different reward. To realize these benefits, we train a policy conditioned on natural-language reward axes and their corresponding scores.

In principle, there are many base RL algorithms that we can extend to the language-conditioned setting with a language-conditioned reward function. We opt to use a particularly simple approach, based on prior work that trains reward-conditioned policies~\cite{chen2021decision,schmidhuber2019reinforcement,kumar2019reward,intelligence2025pi}. Because we would like our policy to be able to optimize all representative axes of preferences rather than just one, we select a comprehensive set of $K_\pi$ preference axes for policy training $L = \{l_k | k=1\dots K_\pi\}$. These can be selected as representative axis descriptions that appear in the reward model training dataset, either manually or through automatic summarization to remove synonym phrases. Then, we condition the policy on all of the axis descriptions $l_k$ and corresponding trajectory rewards $r_\phi(\tau | l_k)$. More formally, our policy training objective is:

\begin{equation}
\label{eq:multi-axis-fm}
\mathcal{L}_{\text{\Acronym-Policy}}(\theta)
=
-\mathbb{E}_{\tau_i \sim \mathcal{D}}
\left[
\sum_{t=1}^{T_i}
\log \pi_\theta\!\left(
\mathbf{a}_i^t
\,\middle|\,
s_i^t, l_1, r_\phi(\tau_i \mid l_1) \ldots l_{K_\pi}, r_\phi(\tau_i \mid l_{K_\pi})
\right)
\right],
\end{equation}
\vspace{-0.5em}

This approach can be used in both an offline and online RL setting. In the former case, this policy optimization step is performed once on an offline preference dataset. In the latter case,  we repeat this process: collecting roll-out data from the latest policy, soliciting preference annotations on this new data, and then updating the policy.

\begin{wrapfigure}{r}{0.52\linewidth}
\begin{minipage}{0.52\textwidth}
\vspace{-0.35cm}
\begin{algorithm}[H]
\captionof{algorithm}{ \Acronym: \MethodName}
\label{alg:fpl}
\begin{algorithmic}
\Require Initial offline dataset $\mathcal{D}$, number of iterations $N$, $\pi_0$ initial policy
\State $\mathcal{P} \gets \emptyset$;  Initialize preference buffer

\For{$n = 1, \dots, N$}
    \State $\mathcal{P} \gets \mathcal{P} \cup \mathcal{P}_n$; Collect $\mathcal{P}_n$ freeform preferences over pairs from $\mathcal{D}$
    \State $r^n_{\phi} \gets$ Train reward model on $\mathcal{P}$ ( Eq.~\ref{eq:multi-axis-bt})
    \State $L_n \gets $ Extract preference axes from $\mathcal{P}$ 
    \State $\pi_n \gets $ Train $\pi_n$ using $r_\phi^n$ on $\mathcal{D}$ (see Eq \ref{eq:multi-axis-fm})
    \If{$n \neq N$}
        \State $\mathcal{T}_n \gets$ Roll out $\pi_{n}$ 
        \State $\mathcal{D} \gets \mathcal{D} \cup \mathcal{T}_n$
    \EndIf
\EndFor
\State \Return final policy $\pi_N$
\end{algorithmic}
\end{algorithm}
\vspace{-0.5cm}
\end{minipage}
\end{wrapfigure}

Notably, the preference axes described by human annotators may evolve over multiple iterations as the policy becomes more capable, and likewise, the set of $K_\pi$ preference axes for policy optimization can also change and expand. This naturally can yield a curriculum where initial preferences focus on initial stages of the task or coarse attributes of behavior, while later preferences can focus on later stages of the task and fine details. The full iterative training process is outlined in Algorithm~\ref{alg:fpl}.

At test-time, \Acronym~ can steer the policy towards high-performing behaviors by varying the target rewards used for conditioning. 
Because the reward model outputs are unbounded, selecting values that both elicit the desired behavior and remain in distribution can be difficult. We therefore standardize rewards per axis over $\mathcal{D}$, yielding a normalized scale for more easily querying the policy at test time.

\section{Related Work}
\textbf{Real-world reinforcement learning on vision language action models}: Performing reinforcement learning in the real world on vision language action models (VLAs) \cite{blackpi0,intelligence2025pi,nvidia2025gr00tn1openfoundation,kim2024openvla,torne2026mem} is challenging for several reasons. VLAs typically use a generative action head \cite{blackpi0,chi2023diffusion}, and a line of prior work studies how to apply RL to such generative policies \cite{ren2025diffusion,xu2026rl,ankile2025imitation}. To make real-world RL more sample-efficient, we build on reward-conditioning \cite{kumar2019reward,chen2021decision}, an off-policy policy-extraction approach that can also be applied offline \cite{kumar2019reward,emmons2021rvs,intelligence2025pi,chen2021decision,schmidhuber2019reinforcement}. However, all prior work conditions on a single scalar return. In particular, \citet{intelligence2025pi} proposes to learn a value function from success/failure and time-to-go signals and conditions the VLA on a binarized advantage. While effective, this signal is sparse, derived from a single task outcome, which limits its effectiveness on long-horizon tasks where successes might be scarce. Other works instead derive the value function from pretrained language models \cite{liang2026robometer,chen2026topreward}, but likewise reduce it to a single scalar. We instead learn the reward model and condition the policy on multiple reward axes simultaneously, leading to denser signal and policy capabilities such as compositionality of behaviors and steerability. Relatedly, multi-objective RL learns policies conditioned on a reward-weight vector over a fixed set of objectives \cite{hwang2024promptable,peschl2021moral,anqi2020morl,Moffaert2013ScalarizedMR}, allowing behaviors to be steered at test time. Unlike these methods, we do not assume predefined axes or rewards, but learn from freeform human preferences.

\textbf{Reinforcement learning from human preferences}: Learning from human preferences sidesteps the difficulty of hand-designing reward functions, and has proven effective for large language models \cite{ouyang2022training,ziegler2019fine,rafailov2023direct}. It is especially valuable in real-world robotics from image observations, where engineering a reward function is notoriously hard. In robot learning, \citet{christianoHumanPref} and much subsequent work \cite{lee21pebble,torne2023breadcrumbs,balsells2023autonomous,hejna2024contrastive,hejna2023inverse,lin2026peel} learn a reward model from binary preferences comparing two robot segments. However, a single overall preference on a long-horizon task is often ambiguous and we instead collect freeform preferences in which annotators specify the axes they wish to judge to reduce the ambiguity. A complementary line of work studies which comparisons to query, using active selection to learn reward functions from fewer preferences \cite{Sadigh2017ActivePL,biyik18prefs}. \citet{bobu2022inducing} decompose reward learning into explicitly taught reward features \cite{basu2018learning, peng2024pragmatic} that are combined into a single reward and \citet{agnihotri2025multi} optimizes a primary objective while enforcing lower bounds on secondary objectives. In contrast, we let annotators specify preference axes directly in natural language and optimize the policy over these.
A separate direction leverages the state-of-the-art VLMs to obtain binary preferences \cite{wang2024rl}, some querying for specific substeps to obtain denser supervision \cite{zhang2024grape}. However, these do not leverage the granularity and multi-dimensionality of rewards as we propose with \Acronym. 

\textbf{Learning from natural language supervision}: Denser human feedback aims to provide a stronger learning signal to the RL process. In the LLM setting, \citet{wu2023fine} trains multiple reward models, each tied to a distinct error category. In contrast, we learn a single language-conditioned reward model from freeform human-specified axes rather than a fixed, predefined category set. Other methods leverage richer feedback through the model's capacity for in-context adaptation from natural language, for example rich human feedback for text-to-image generation \cite{liang2024rich} and open-ended text optimization via pairwise comparison \cite{lee2025feedback}. While in-context adaptation is appealing, current VLAs cannot react to language feedback in-context in the same way making such methods inapplicable to robot learning. Prior robotics work has used natural language to correct robot plans \cite{sharma2022correcting}, and to infer preference-conditioned state abstractions that identify task-relevant features for reward inference \cite{peng2024preference}. In contrast, we use language not merely as corrective feedback, but as an explicit interface for defining the axes along which preferences are elicited and rewards are learned. Finally, \citet{hirota2025active} learn a shared latent space between comparative language and trajectories, and \citet{hwang2026causally} treat freeform rationales attached to preferences as projection axes in an embedding space. Both must first learn a dedicated space in which language and trajectories are aligned, and both use that language to sharpen a single scalar reward, whereas we learn a open-ended language conditioned reward model and train the policy with unmerged axes, letting it compose behaviors absent from the demonstration data.

\section{Analysis \& Experimental Evaluation}

Our experimental section is designed to answer the following questions: (a) Does \Acronym learn effective policies through freeform human preferences? (b) Do policies learned through \Acronym exhibit compositionality of behaviors unseen in the data? (c) Does \Acronym exhibit steerability of rewards at test time? (d) Does \Acronym learn denser reward functions for long-horizon tasks? 

\subsection{Experimental Setup}

In order to empirically respond to the questions above, we consider four real-world manipulation tasks and two simulation tasks. 

\textbf{Real-world.} \emph{Put cube into target bowl} is a diagnostic task for steerability, where the policy is reward-conditioned to place the cube into one of three bowls. \emph{Fold shorts} tests deformable-object manipulation, requiring the robot to fold shorts in three folds while optimizing speed and alignment. \emph{Plate toast} tests dexterous tool use, requiring the robot to transfer toast from a tray to a plate while using the spatula smoothly. \emph{Set up the table} is a long-horizon task in which the robot places two plates, cutlery, and a cup, testing task completion, formality, and carefulness. All real-world tasks use the DROID setup~\cite{khazatsky24droid}, with two camera views as observations and joint-velocity control for a Franka robot. The tasks are shown in Figure~\ref{fig:taskoverview}; further details are in Appendix~\ref{appdx:realworlddetails}.

\textbf{Simulation.} We use two Robomimic-based tasks~\cite{robomimic2021}. \emph{Object rearrangement} requires placing two objects into the correct containers and in the correct order. \emph{Bimodal square} tests compositionality, the target behavior is to place the nut on the right peg quickly, while the initial dataset contains fast left-peg trajectories and slow right-peg trajectories. We also evaluate \emph{bimodal square (inverted)} to test whether the same policy can be steered at test time to place the nut on the left peg. More details can be found in Appendix \ref{appdx:simdetails}.

All tasks start from offline demonstrations with varying quality and strategies. We report mean and standard error, using 20 rollouts per real-world method and three seeds in simulation (see Apdx. \ref{appdx:expdetails}).

\subsection{Experimental Evaluation}
\label{sec:experimentaleval}
\begin{figure}[t]
    \centering
    \includegraphics[width=\linewidth]{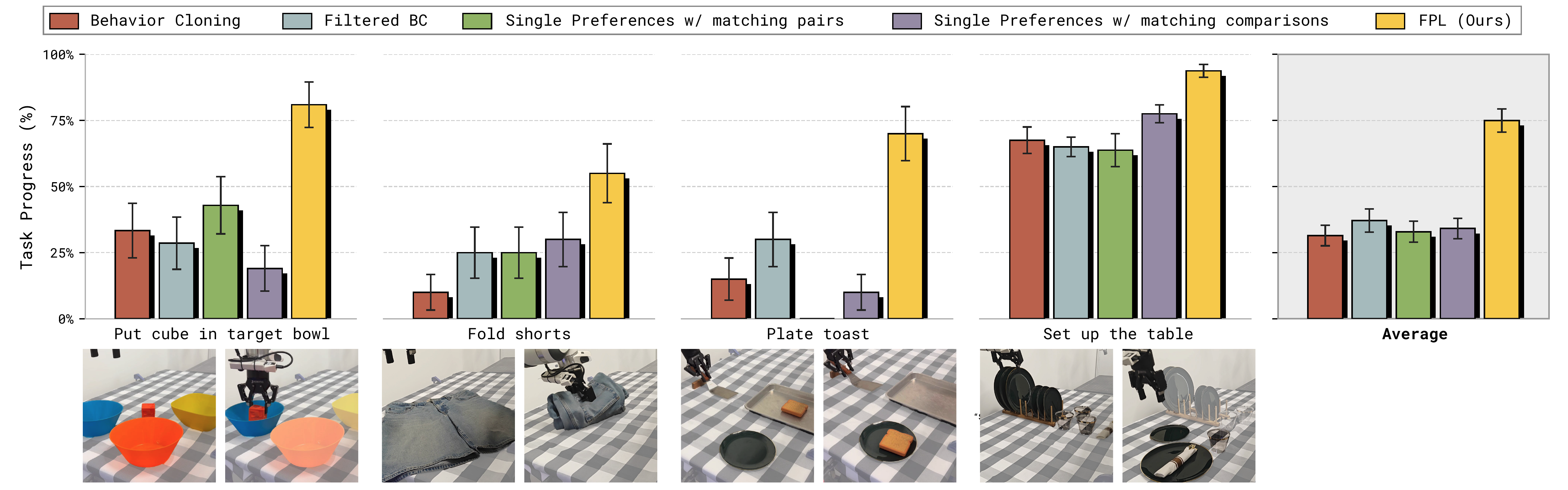}
    \vspace{-1.5em}
    \caption{\footnotesize Baseline performance on real-world tasks.  \Acronym learns more performant policies across all tasks, with an average improvement of 38 percentage points over the second-best baseline. The signal from binary preferences is too ambiguous to learn performant policies. The sparse success signal is too weak to provide the necessary supervision to solve the tasks successfully. 
    }
    \label{fig:realworldresults}
    \vspace{-1.5em}
\end{figure}

\textbf{Comparisons.} We compare \Acronym against five baselines. \emph{Single Preferences} learns from standard pairwise preferences over a single ``overall quality" axis \cite{christianoHumanPref}, using the same reward-conditioned policy extraction as \Acronym for a fair comparison. \emph{Advantage Conditioning} following \citet{intelligence2025pi}, we train a value function from success signal and time-to-go supervision, and condition the policy on the resulting advantage. \emph{Weighted Regression} uses the multi-dimensional reward model learned by \Acronym, but extracts the policy with weighted regression using the average of rewards across the axes $L$ \cite{peng2019advantage}. \emph{Filtered BC} trains on the offline dataset together with successful policy rollouts, providing a sparse reward policy-extraction baseline \cite{emmons2021rvs}. \emph{BC} trains with imitation learning over the original offline dataset without reward learning or iterative improvement \cite{levine2016end}.

\begin{table}
\centering
\caption{\footnotesize Simulation results comparing \Acronym against baselines across simulation environments. Multi-dimensional preferences as leveraged by \Acronym provide the best supervision. \Acronym can successfully solve with the same policy the \emph{bimodal square} and \emph{bimodal square inverted} benchmarks by changing the reward conditioning at test-time.}
\label{tab:sim-results}
\small
\setlength{\tabcolsep}{6pt}
\begin{tabular}{lccc}
\hline
 & Object rearrangement & Bimodal square & Bimodal square \\ 
Method & (success) & (throughput) & inverted (throughput) \\
\hline
BC \cite{levine2016end}      & $0.04 \pm 0.02$ & $0.71 \pm 0.07$ & $0.25 \pm 0.14$ \\ 
Filtered BC \cite{emmons2021rvs} & $0.09 \pm 0.05$ & $0.71 \pm 0.02$ & $0.00 \pm 0.00$ \\

Weighted Regression \cite{peng2019advantage}  & $0.18 \pm 0.08$ & $0.83 \pm 0.08$ & $0.09 \pm 0.06$ \\ 
Advantage Conditioning \cite{intelligence2025pi}  & $0.17 \pm 0.06$  & $0.75 \pm 0.12$ & $0.08 \pm 0.04$ \\ 
Single Preferences (match pairs)    \cite{christianoHumanPref}    & $0.73 \pm 0.04$ & $0.67 \pm 0.08$ & $0.78 \pm 0.01$ \\ 
Single Preferences (match comparisons)    & $0.79 \pm 0.06$ & $0.73 \pm 0.05$ & $0.74 \pm 0.05$ \\
\Acronym (Ours)           & $\mathbf{0.84 \pm 0.09}$ & $\mathbf{1.19 \pm 0.01}$ & $\mathbf{1.24 \pm 0.01}$ \\
\hline
\end{tabular}
\vspace{-1.5em}
\end{table}

\textbf{\Acronym learns performant policies from freeform human preferences}. In Figure \ref{fig:realworldresults}, we show that \Acronym outperforms all baselines in the real-world, improving by 38 overall percentage points over the next best method. We find that single overall preferences are often ambiguous in long-horizon tasks. For example, in \emph{plate toast}, one trajectory may use the gripper fingers instead of the spatula, making it unhygienic, while another may use the spatula but drop the toast. In such cases deciding which trajectory is ``better overall" is difficult and leads to noisy supervision. Sparse success/failure rewards are also insufficient: in \emph{setup table}, the robot must complete several sequential subtasks, so rewarding only perfect executions provides too little signal, while rewarding imperfect executions can reinforce undesirable behavior. In practice, sparse-reward baselines learned to place the items in the correct locations but did not learn to place the items carefully and often dropped the plates instead of placing them with care. In contrast, \Acronym provides axis-specific supervision, allowing the policy to improve both the task completion and qualitative aspects of behavior.

\begin{figure}[t]
    \begin{minipage}[t]{0.38\linewidth}
        \centering
        \includegraphics[width=\linewidth]{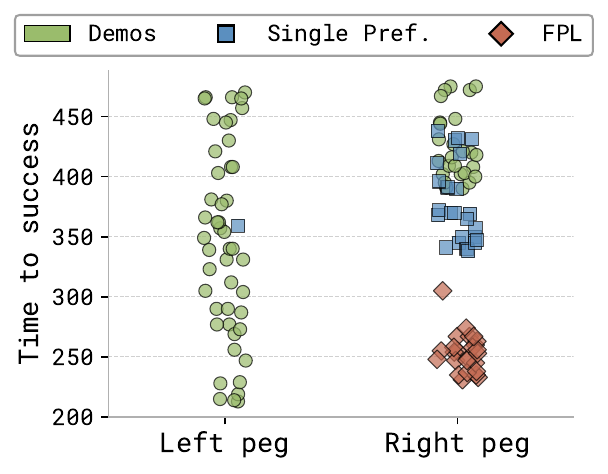}
        \captionof{figure}{\footnotesize Time to success per peg and per baseline. \Acronym achieves fast trajectories on the right peg despite the demo data not including fast data on the right peg, only on the left peg. \Acronym exhibits compositionality of behaviors while baselines do not.}
        \label{fig:compositionality}
    \end{minipage}\hfill
    \begin{minipage}[t]{0.55\linewidth}
        \centering
        \includegraphics[width=\linewidth]{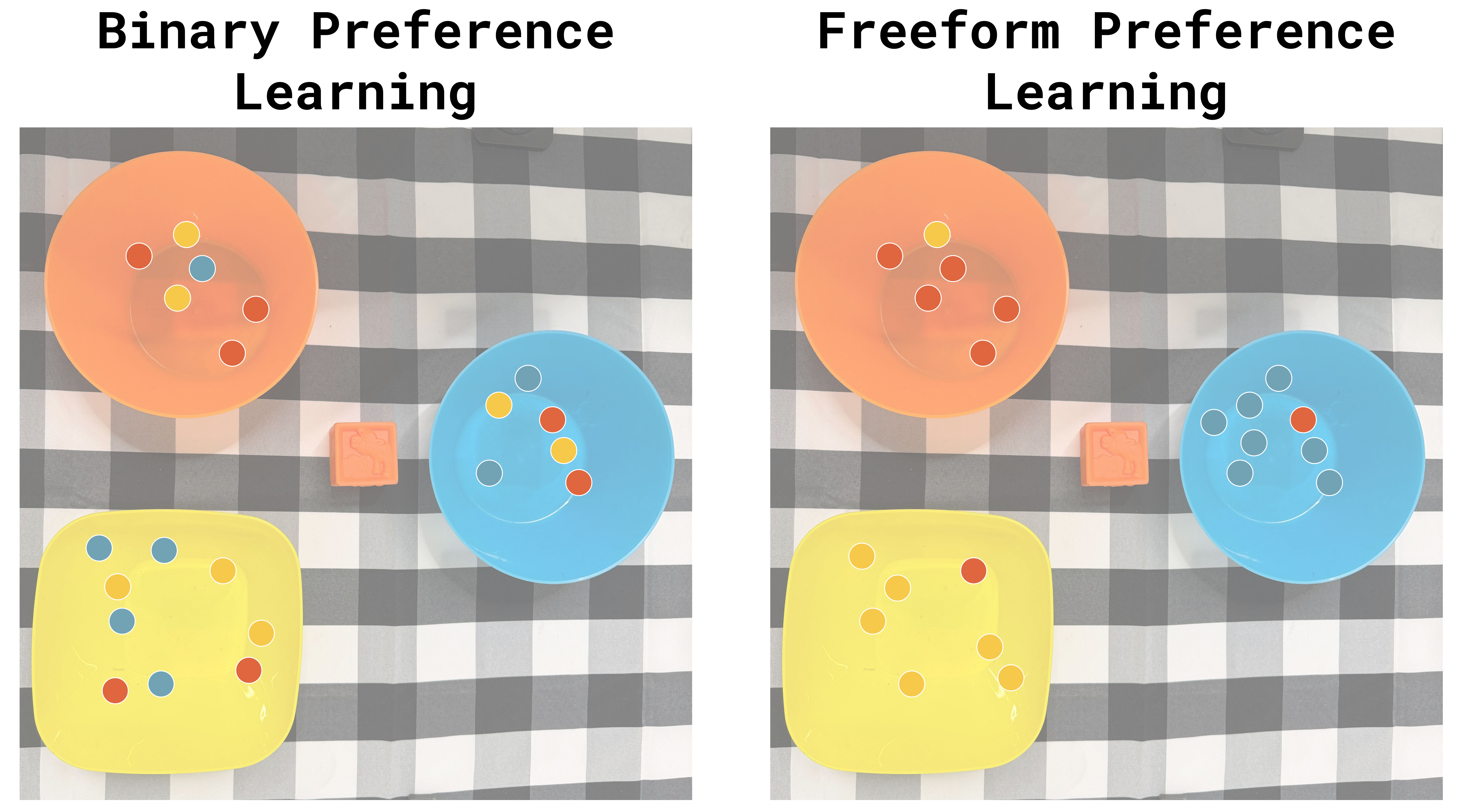}
        \captionof{figure}{\footnotesize Each dot represents the block's final position, with the color corresponding to the target bowl's color. With \Acronym we can steer the policy at test-time by prompting a different reward to optimize while the baselines cannot.}
        \label{fig:steerability}
    \end{minipage}
\end{figure}

In simulation, \Acronym outperforms all baselines, as shown in Table \ref{tab:sim-results}. The gap is particularly clear in the \emph{object rearrangement} task where the long-horizon and noisiness of the offline dataset has few complete successes.  In this setting, success/failure signal is too sparse to correctly learn a successful policy. By contrast, preference-based methods provide denser signal. But, single overall preferences still collapse multiple behaviors into one ambiguous signal. As a result the policies learned from single preferences achieve the correct arrangement of objects but fail since they often drop objects from too high above the target. \Acronym, however, uses multi-dimensional preferences to capture the different important axes for the task separately, leading to successful policies.

\textbf{\Acronym exhibits compositionality of behaviors}. We study this in the \emph{bimodal square} simulation environment, where the goal is to place the nut on the right peg. As shown in Figure \ref{fig:compositionality}, the offline dataset has both fast and slow demonstrations for the left peg but \emph{only} slow demonstrations for the right peg. \Acronym achieves faster right peg placements than those seen in the training data, while the single preference baseline does not improve beyond the demonstrated behaviors. This shows that \Acronym can compose behaviors through the preference axes, in this case the target placement at faster speed, whereas the baselines cannot (see Table \ref{tab:bimodal-square-detailed}). We attribute this compositionality to multi-dimensional reward learning together with reward-conditioned policy extraction.

\textbf{\Acronym exhibits test-time steerability}. We evaluate steerability using the \emph{inverted} version of \emph{bimodal square}, where the same trained policy is conditioned to place the nut on the left peg instead of the right peg. As we show in the \emph{inverted} columns of Table \ref{tab:sim-results} and Figure \ref{fig:steerability}, \Acronym is the only method that achieves high performance on both the original and the inverted tasks with the same policy. This is enabled by reward-conditioned policy extraction on multi-dimensional rewards: because \Acronym trains on trajectories across the replay buffer without filtering, the policy observes both high- and low-scoring behaviors along each axis and can be steered at test time by changing the target reward conditioning.

\subsection{Qualitative Analysis}

\begin{figure}[t]
    \centering
    \includegraphics[width=\linewidth]{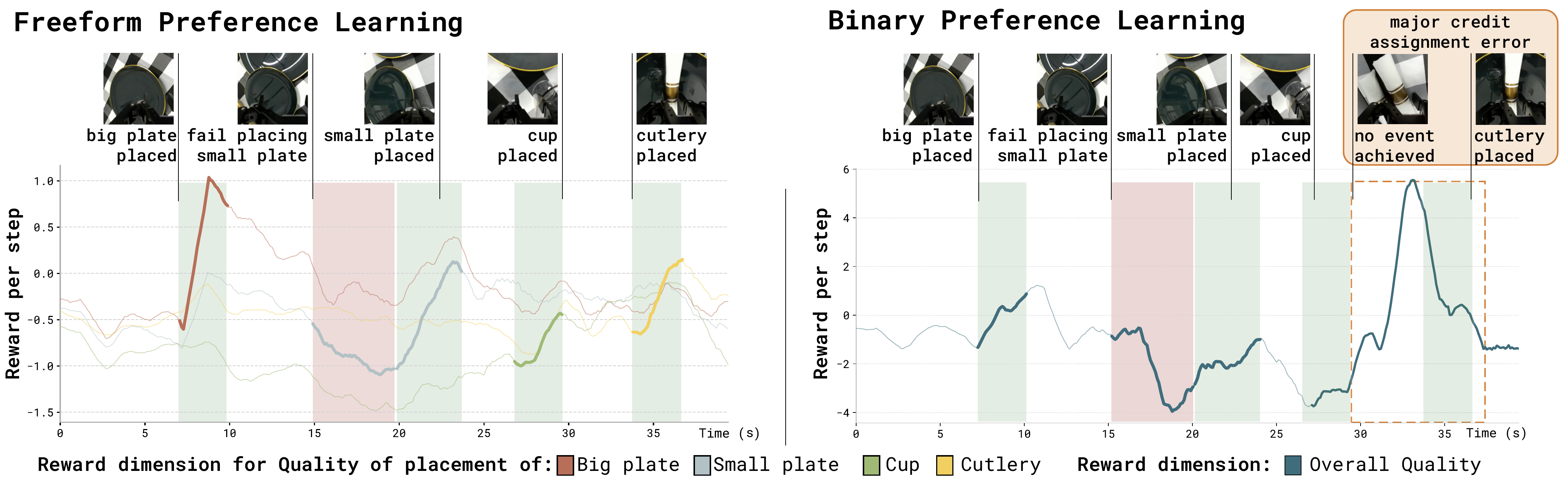}
    \vspace{-1.5em}
    \caption{\footnotesize Qualitative reward analysis on \emph{setup table}. We compare rewards learned from freeform preferences with \Acronym (\emph{left}) and from binary preferences (\emph{right}) on a representative rollout. For visualization, we show a subset of \Acronym’s reward axes. The axis-conditioned rewards learned by \Acronym are temporally localized around their corresponding subtask events, while the binary-preference reward collapses behavior into a single score and produces a large spike near the end of the trajectory despite no major subtask being completed.}
    \label{fig:rewardanalysis}
    \vspace{-1.5em}
\end{figure}

\textbf{\Acronym qualitatively produces denser reward signals on long-horizon tasks without explicit task segmentation}. In Figure \ref{fig:rewardanalysis}, we qualitatively compare reward models learned with \Acronym and binary preference feedback on an example rollout from the \emph{setup table} task. Although neither model is trained with explicit subtask boundaries, the reward learned with \Acronym temporally localized the corresponding events such as placing the big plate, small plate, cup, and cutlery. This makes the learned reward more interpretable and suggests that freeform, axis-specific preferences can provide a denser signal for long-horizon tasks. In contrast, the binary-preference reward produces a large reward spike near the end of the episode, despite no major subtask being completed at that point showing then an error in the credit assignment.

\textbf{Freeform preferences axes naturally change throughout iterations}. With the policy performance improving at each iteration, the annotators change from more coarse feedback to more concrete and perfectionist feedback. In Figure \ref{fig:axis-trends}, we observe that in the early iterations of \emph{fold shorts} task, the annotator focuses on whether each fold happens at all, and later ones, once folding is reliable, whether there are wrinkles and final alignment are perfected. 

\textbf{Freeform preferences produced diverse preference axes}. Allowing annotators to specify preference axes in natural language yields a diverse set of labels. In the \emph{plate toast} task, we obtained 295 trajectory-pair comparisons and over 1477 axis-level comparisons spanning 41 distinct labels (see Figure \ref{fig:prefdetails} in Appendix \ref{appdx:preference-collection}). This suggests the annotators naturally use a wide range of criteria when evaluating robot behavior, motivating the need for preference learning methods that preserve this structure rather than collapse feedback into a single overall score.

\textbf{Freeform preferences reduce annotation time per label.} As shown in Figure~\ref{fig:annotation_time}, collecting freeform preferences is approximately 50\% faster per label than collecting single binary preferences. Because annotators provide multiple axis-specific judgments for each trajectory pair, the cost of viewing the videos is shared across several labels, which reduces the annotation overhead.
\section{Limitations \& Conclusion}

To conclude, we introduced \Acronym, a method for learning robot policies from freeform human preferences. By collecting preferences with natural-language axes, \Acronym provides denser and less ambiguous supervision than single binary preferences. Across four real-world tasks and two simulation settings, \Acronym outperforms the baselines. We show that preserving the multi-dimensional structure of human feedback enables compositionality of behaviors and test-time steerability of the learned policy as well as qualitatively learn reward models with better credit assignment. 

Several limitations remain. Preference learning requires collecting human preferences, which is more expensive than fully unsupervised approaches. Reward-conditioned policy learning requires selecting appropriate reward values at test time, automating this selection is an important direction for future work. Finally, the current policy is conditioned on a fixed set of preference axes, and extending this method to handle variable axes is a promising direction as VLAs become more capable.


\clearpage
\makeatletter
\section*{Acknowledgments}

 We thank Lars Ankile and Moo Jin Kim for their support with the real world experiments. We thank Alex Swerdlow, Aneesh Muppidi, Yuejiang Liu and the members of the Stanford IRIS lab for their insightful discussions and feedback on the paper drafts. This work was supported in part by the ONR grant N00014-22-1-2621, the Robotics and AI Institute and, the Stanford Institute for Human-Centered AI. C. Finn is a CIFAR fellow.

\makeatletter
\section*{Contributions}

The project was started and designed by MT and CF. MT designed the overall algorithm. MT implemented the reward model and conditioning of the policy. MT designed and implemented the simulation benchmarks and baselines. AM, AB and MT worked on the real world experiments including data and preference collection as well as fine-tuning of the real world policies. CF provided advice throughout the project. MT and CF worked on writing, illustrations, blog post and the video.


\bibliography{example}  

\clearpage

\section*{Appendix}

\section{Experimental details}
\label{appdx:expdetails}

In this section, we will go over the details of the simulation \ref{appdx:simdetails} and real-world \ref{appdx:realworlddetails} experiments presented in the paper.

\subsection{Real-World Tasks}
\label{appdx:realworlddetails}
\begin{figure}[h!]
    \centering
    \includegraphics[width=\linewidth]{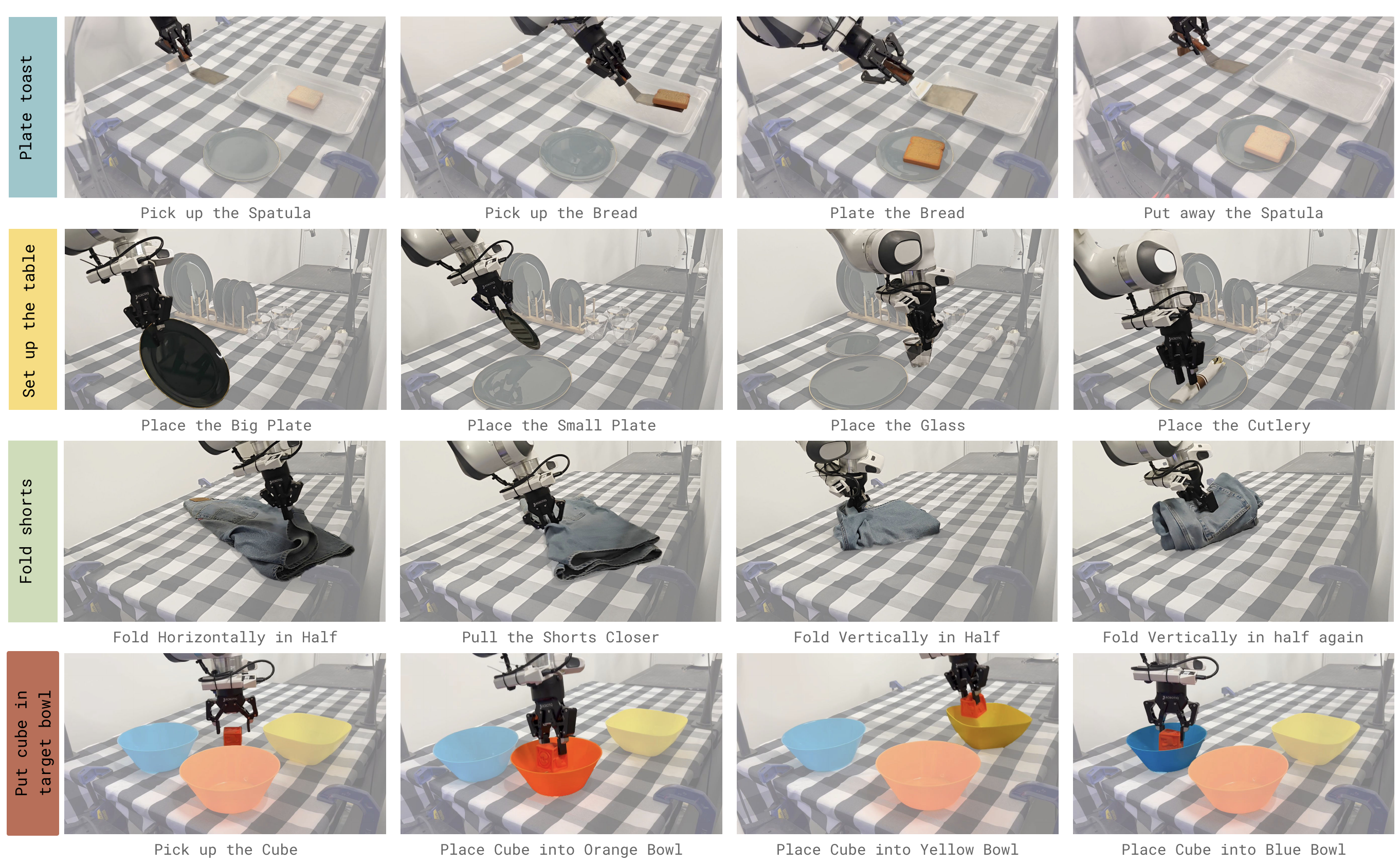}
    \caption{ \footnotesize Real-world task suite. Each row shows key stages of the corresponding task.}
    \label{fig:taskoverview}
\end{figure}

All four real-world tasks (see Fig. \ref{fig:taskoverview}) use the DROID setup~\cite{khazatsky2024droid} on a Franka robot, with two camera views (wrist and third person) as observations and joint-velocity control. Each task starts from offline demonstrations of varying quality and strategy. The per-task descriptions and preference axes are described below.

\paragraph{Plate Toast.}
The robot is tasked with plating a piece of toast, either by scooping it with a spatula or picking it up directly with the gripper from a tray and placing it onto a plate. The initial position and orientation of the toast on the tray vary between episodes, as does the position of the spatula. Unlike the structured tasks, the preference axes were not fixed in advance; instead, the annotators freely specified the dimensions most relevant to them. The collected axes include speed, smoothness of motion, cleanliness, quality of plating, and whether the robot damaged the toast.

\paragraph{Setup Table.}
Here the robot lays out a full place setting on a dining table: a large plate, a small plate, a cup, and cutlery. The small plate can be placed to the left of the main plate (most formal), on top of it (medium formal), or to the right of it (least formal). The cup is placed on the opposite side of the small plate, except when the small plate is on the main plate, in which case the cup goes to the left. The cutlery is placed on the main plate in the center. The initial placement of each item on the table varies between episodes. Preferences are collected across the placement quality of each item (large plate, small plate, cup, and cutlery), formality of the resulting setup, smoothness and carefulness of motion, speed, and damage to the environment.

\paragraph{Fold Shorts.}
The robot folds a pair of shorts lying flat on a surface by performing three successive folds, each folding the garment in half. The initial configuration of the shorts, including its position and orientation on the surface, varies between episodes. Preferences are collected along the following axes: quality and wrinkle level of each fold (1st, 2nd, 3rd), alignment of the final folded garment, speed of execution, smoothness of motion, and any damage caused to the environment.

\paragraph{Place Cube into Bowl.}
In this task, the robot places a cube in one of three bowls (blue, orange, or yellow) on the table. The initial position of the cube and the bowls varies between episodes. The task is designed to evaluate the steerability of the learned policy at test time. Preferences are collected along the following axes: target bowl placement (blue, orange, yellow), speed of execution, and smoothness of motion.

\paragraph{Dataset Statistics.}
Table~\ref{tab:task-details} details the dataset statistics for real-world task.
\begin{table}[H]
\centering
\caption{ \footnotesize Dataset statistics for each task, including the number of offline demonstrations, total preference pairs, and total rollouts used for preference learning.}
\label{tab:task-details}
\small
\setlength{\tabcolsep}{6pt}
\begin{tabular}{lccccc}
\hline
Task & Offline & Num & Num Rollouts & Preference & Num Preferences (single  \\
 & Dataset & Preferences &  & axes &matching baseline only) \\
\hline
Plate Toast & 148 & 295 & 128 & 5 & 1492\\
Setup Table  & 100 & 325 & 270 & 5 & 1625\\
Fold Shorts & 138 & 239 & 130 & 10 & 1336*\\
Place Cube into Bowl & 0 & 250 & 400 & 3 & 800\\
\hline
\end{tabular}
\end{table}

(*) to match the number of labels would have been prohibitively costly, we still allocate more time to collect those labels than for \Acronym for a fair comparison see Section \ref{appdx:preference-collection} on more analysis on the cost to collect single preferences against freeform preferences.

\subsection{Simulation Tasks}
\label{appdx:simdetails}
We propose two simulation tasks and one additional variation (see Fig. \ref{fig:simtaskoverview}), which are extensions from the robomimic original tasks ~\cite{robomimic2021,zhu2020robosuite}. Note that the simulation benchmarks will be available with the open-sourced version of the code. We describe them in further details below:

\begin{figure}[h!]
    \centering
    \includegraphics[width=\linewidth]{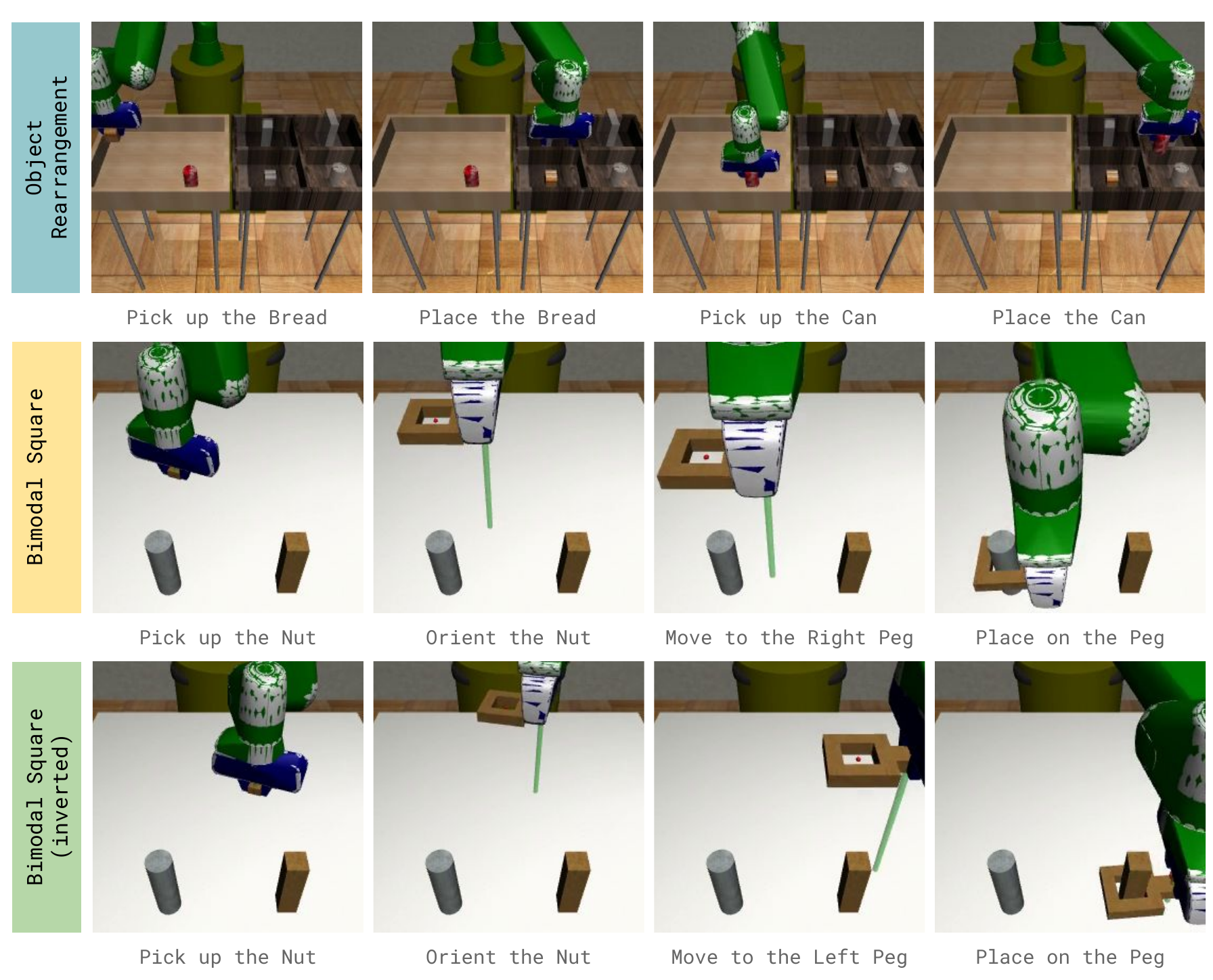}
    \caption{Simulation task suite. Each row shows key stages of the corresponding task.}
    \label{fig:simtaskoverview}
\end{figure}

\paragraph{Object Rearrangement.} This task is a modification of the pick and place task from \citet{zhu2020robosuite}. Two objects are present in the scene, the \emph{bread} and the \emph{can}. The goal is to place them in the correct box, in the correct order (first \emph{bread} then \emph{can}), and without dropping them. The offline dataset contains a mix of behaviors, data that places in the wrong order or that drops the objects as well as perfect demonstrations.

\paragraph{Bimodal Square.} This task is a modification of the square task from \citet{zhu2020robosuite}. The goal of this task is to evaluate the compositionality of the policies. The task is considered successful if the nut is placed on the right peg, and we also keep track of the time to success. The composition of the offline dataset as shown in \ref{fig:compositionality} is made of fast and slow trajectories for placing the nut on the left peg, but \emph{only} slow trajectories for placing the nut on the right peg. 
\\

\emph{Bimodal Square (inverted).} This is an extension of the \emph{Bimodal square task}. The goal of this task is to evaluate the test-time steerability of the policies. The goal becomes to place the nut on the left peg instead of the right peg (as the success specified at training time). It is possible to steer the policy at test time for \Acronym as show in Section \ref{sec:experimentaleval} but not possible for the rest of the policies. 

\paragraph{Dataset Statistics.} In Table \ref{tab:task-sim-details}, we provide further details on the dataset and preference axes for the simulation tasks. 

\begin{table}[h!]
\centering
\caption{ \footnotesize Dataset statistics for each task, including the number of offline demonstrations, total preference pairs.}
\label{tab:task-sim-details}
\small
\setlength{\tabcolsep}{6pt}
\begin{tabular}{lcccc}
\hline
Task & Offline  & Num  & Preference axes & Num Preferences (single\\
 & Dataset & Preferences &  & matching baseline only)\\
\hline
Object  & 300 & 70 & order, bread placed, can placed & 350 \\
rearrangement&  &  &  bread dropped, can dropped &  \\
\\
Bimodal square & 200 & 100 & speed, peg & 200\\
\hline
\end{tabular}
\end{table}

\section{Preference Collection}
\label{appdx:preference-collection}

\begin{figure}[ht]
    \centering
    \begin{subfigure}[b]{0.62\linewidth}
        \centering
        \includegraphics[width=\linewidth]{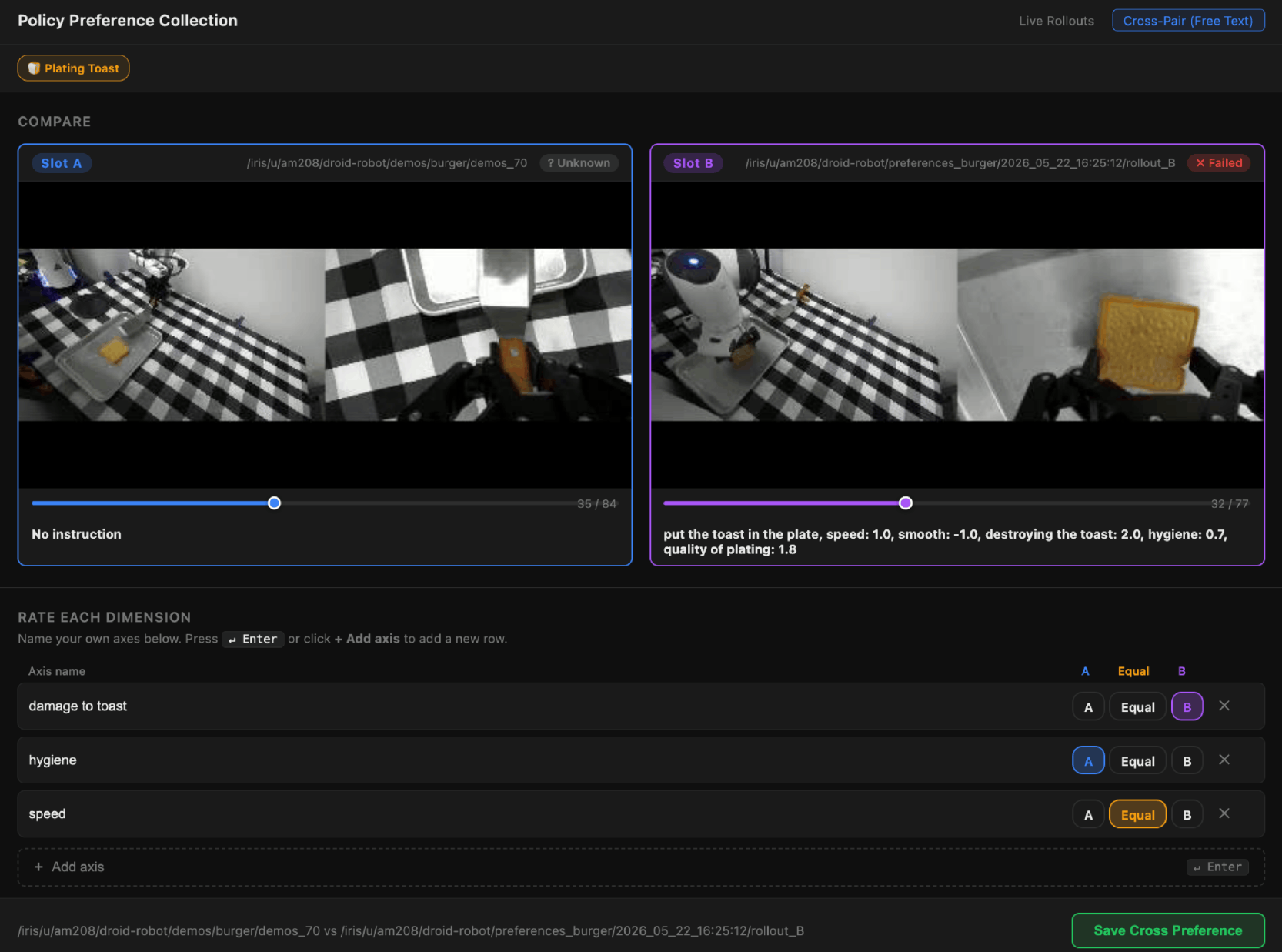}
        \caption{Data collection interface}
        \label{fig:interface}
    \end{subfigure}
    \hfill
    \begin{subfigure}[b]{0.36\linewidth}
        \centering
        \includegraphics[width=\linewidth]{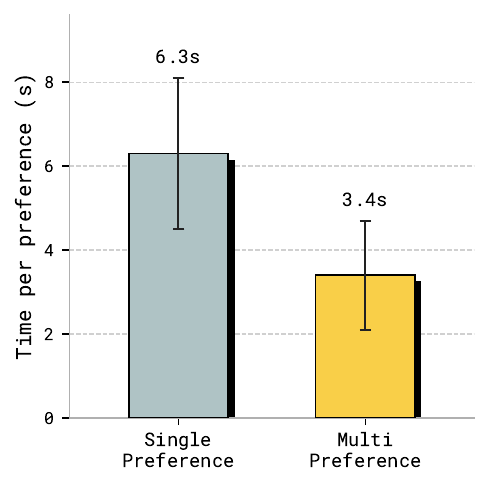}
        \caption{Annotation time}
        \label{fig:annotation_time}
    \end{subfigure}
    \caption{\footnotesize \textbf{(\subref{fig:interface}) Data collection interface.} Annotators are shown two side-by-side robot rollout videos and asked to indicate which rollout is preferred along each axis or indicate equivalence. For open-ended tasks, annotators first specify their own preference axes in free-form text. \textbf{(\subref{fig:annotation_time}) Annotation cost.} Average wall-clock time to collect a single preference label. Under single-preference annotation, each pair of rollout videos yields one labeled axis, whereas multi-preference annotation elicits several axes from the same pair. Error bars denote one standard deviation across annotation sessions.}
    \label{fig:datacollection}
\end{figure}

Preferences are collected through a custom web interface in which annotators are presented with two side-by-side video recordings of robot rollouts and asked to indicate which rollout is preferred along each axis or indicate equivalence. For some tasks (Setup Table, Fold Shorts, and Place Cube into Bowl), the axes are predefined and vary by task. For the others, like the Plate Toast task, annotators additionally specify their own preference axes in free-form text prior to providing their preferences; this allows for richer feedback along dimensions of quality that would otherwise go unspecified. Figure~\ref{fig:interface} shows the data collection interface.

In Figure \ref{fig:annotation_time}, we show how collecting freeform preferences is cheaper in time than collecting single preferences when looking at the time spent per label. The annotator can provide multiple labels instead of only one when watching a single pair of videos of two trajectories, while for single preferences they can only provide one label, therefore the overhead cost is less compensated. The speedup observed in our experiments is of around 1.85 times faster to provide freeform preferences than single preferences. 

\begin{figure}[ht]
    \centering
    \includegraphics[width=0.6\linewidth]{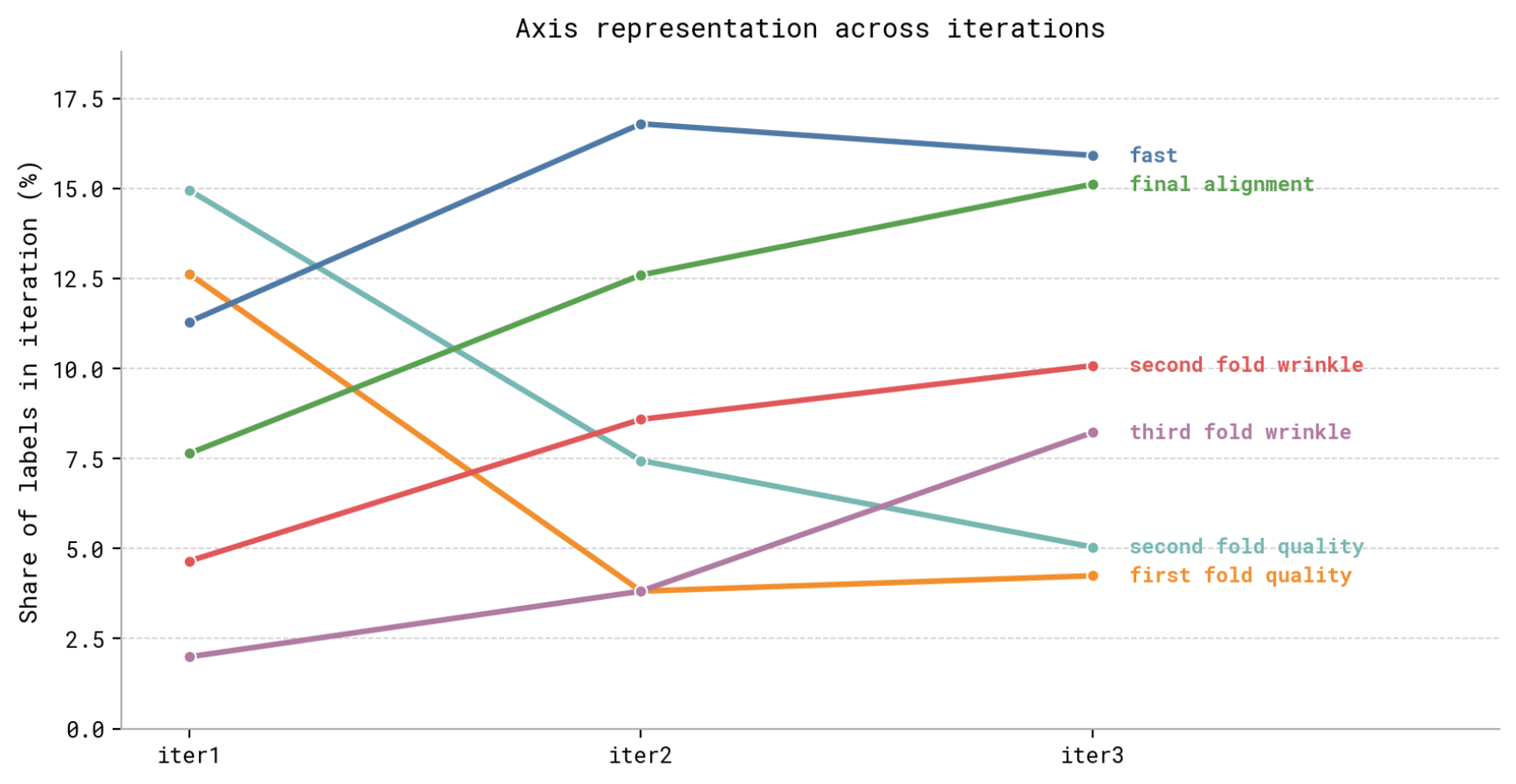}
    \caption{\footnotesize Share of freeform preference labels assigned to each criterion across iterations on the \emph{fold shorts} task. As the policy improves, annotator attention shifts away from coarse fold-quality axes toward finer wrinkle and final-alignment criteria; the evaluation changes naturally without any manual redefinition of the reward.}
    \label{fig:axis-trends}
\end{figure}

\Acronym provides flexibility in the axes of preferences. This flexibility becomes useful as the iterations of the algorithm advance and the policy performance improves. In Figure~\ref{fig:axis-trends} which shows the per-iteration label distribution on the iterative \emph{fold shorts} task, we observe that early iterations focus on whether each fold happens at all, and later ones, once folding is reliable, whether there are wrinkles and final alignment. Because the annotators write their own axes instead of scoring a fixed rubric, the criteria change on their own as the policy gets better, and each round ends up targeting whatever the policy is currently getting wrong, making the preferences by \Acronym more targeted.

\section{Implementation Details}
In this section we give further implementation details for \Acronym. 

\subsection{Reward-Model Learning}

In order to leverage the pretraining of VLMs, we use the pretrained Qwen 3.5 VL 4B \cite{bai2023qwen} model as the backbone for our real-world reward model. As presented in Fig. \ref{fig:method}, the reward model takes as input the question ``What is the score for $l_k$?" where $l_k$ gets substituted in natural language for the name of the preference axis provided in the freeform preferences. After the prompt, we pass the images to the VLM, both first-person and third-person, for the whole sequence. Instead of passing the complete trajectory as input we stride the sequence. We freeze the vision encoder, and we full fine-tune the rest for more efficient training. We maintain a causal mask so all tokens can attend to the text and the past images, but not to the future images. We do a single forward pass across the whole sequence, and we take the final vision separator token for the second image of the pair per timestep, decode it with a single linear layer projection into a score. The final score is computed as the sum of scores across all timestep see Eq \ref{eq:trajectory-reward}. When choosing the reward model, we pick the one with the highest validation accuracy in the first 500 epochs. See below for further details:

\begin{table}[H]
\centering
\caption{\footnotesize Hyperparameters for the reward model training}
\label{tab:hyperparams-reward}
\small
\setlength{\tabcolsep}{6pt}
\begin{tabular}{lccccc}
\hline
Task & Sequence Length & Stride Length & Batch Size & LR & Img Size\\
\hline
Plate Toast & 20 & 20 & 32 & 1e-5 & 128x128x3\\
Setup Table  & 20 & 20 & 32 & 1e-5 & 128x128x3\\
Fold Shorts & 20 & 60 & 32 & 1e-5 & 128x128x3\\
Put cube in target bowl & 20 & 10 & 32 & 1e-5 &128x128x3\\
\hline
\end{tabular}
\end{table}

Another detail is that we allow annotators to specify that two trajectories are equivalent in some axis, in those cases we drop the label.

\subsection{Policy Learning}
\emph{Reward-Conditioning}. In order to determine the reward conditioning, we run inference on the whole dataset $\mathcal{D}$ with the reward model on the selected axes $L$. We compute the statistics of the reward and obtain mean $\mu$ and standard deviation $\sigma$. Then, we standardize the reward $r_{cond}=\frac{r-\mu}{\sigma}$. This makes the rewards bounded and easy to specify at test-time without additional search. The reward conditioning is passed to the policy $\pi$ for trajectory $\tau$ through text by passing $l_k : \textit{round} \biggl( \frac{r_\phi(l_k, \tau) - \mu}{\sigma} \biggr) $ where we pass the scores round up to 1 decimal place.  

We use the $\pi_{0.5}$ \cite{pi05} open source weights as the VLA pretraining, which uses a flow matching action head with action chunk size 16 and rollout 8 actions. We do full finetuning of the policies over 30000 steps with a batch size of 32. The policy obtains the wrist and the third person camera views as input as well as the task prompt with the conditioning as through text described above. Thereafter an example prompt for the VLA will be the following:

\begin{PromptBox}{Fold Shorts}
fold the shorts, Quality of 1st fold: 2.2, Wrinkle of 1st fold: 2.1, Quality of 2nd fold: 0.7, Wrinkle of 2nd fold: 0.8, Quality of 3rd fold: -0.5, Wrinkle of 3rd fold: -0.5, Alignment of final fold: 0.1, Fast: 1.2, Smooth: 1.2, Damage to environment: -0.1
\end{PromptBox}

\begin{PromptBox}{Plate Toast}
put the toast in the plate, speed: 0.5, smooth: -0.8, destroying the toast: -1.5, hygiene: -2.1, quality of plating: -0.2
\end{PromptBox}

\begin{PromptBox}{Setup Table}
set up the table, Quality of placement of big plate: -0.8, Quality of placement of small plate: -0.1, Quality of placement of cup: 0.1, Quality of placement of cutlery: -0.9, Formality of setup: -0.1
\end{PromptBox}

\begin{PromptBox}{Put cube in target bowl}
put the cube in the bowl, Blue bowl: -0.5, Orange bowl: 2.0, Yellow bowl: -0.3
\end{PromptBox}

\subsection{Simulation setup}

In simulation, we prioritize speed of iteration, thereafter the environments and policies work directly from the state space. In that case, we don't use any of the pretrained models and train everything from scratch. For the reward model, instead of taking language as command, we have $K$ different heads, one per preference axis, and we predict all of the axes at the same time. The policy remains a flow-matching policy based on the diffusion transformer code from \citet{chi2023diffusion}. Instead of taking the conditioning as input through text, we encode the input as a float vector and pass it to the policy directly. We observed that for the simulation experiments not to overfit to this input conditioning vector, we had to add some noise of 0.2 after standardization of the rewards. This makes it more robust; however, note that this is not necessary in the real world and is only needed here because we train a smaller network from scratch, which is more prone to overfitting.

\subsection{Baselines}
We compare \Acronym~ against six baselines that ablate key components of our method.

\textbf{Behaviour Cloning (BC)} trains a policy via supervised imitation learning on all offline demonstrations $\mathcal{D}$. Preference data are not used, and the policy does not receive reward conditioning.

\textbf{Filtered BC} extends vanilla BC by augmenting the offline dataset $\mathcal{D}$ with rollouts from the preference dataset that were labeled successful. As with BC, no reward conditioning is used; this baseline isolates the benefit of additional on-policy data from the benefit of preference supervision.

\textbf{Single Preference with matching pairs} follows the same reward-conditioned policy extraction as \Acronym~ (Eq.~\ref{eq:multi-axis-fm}), but replaces the multi-dimensional reward with a single scalar reward trained on a user-specified ``overall quality'' axis. The reward model is trained via the Bradley--Terry objective (Eq.~\ref{eq:bt-states-loss}) on a single axis of feedback. This baseline isolates the benefit of multi-axis feedback from the benefit of reward conditioning alone. This baseline uses as many pairs of trajectories compared as \Acronym.

\textbf{Single Preference with matching comparisons} is exactly the same as the \emph{Single Preference} baseline described above, but instead of matching the number of pairs compared, we match the number of comparisons that the human provides. For \emph{Single Preference} we train with one label per comparison, while for \Acronym, we train on $K$ labels per comparison. To demonstrate that the benefit of \Acronym comes from the detail of the preferences and not from a difference in the amount of labels we collect $K$ times more comparisons for this baseline in order to have a fairer comparison. See Table \ref{tab:task-details} and \ref{tab:task-sim-details}, where we explicitly show the number of labels and comparisons collected for this baseline versus the rest.
 
\textbf{Advantage Conditioning} is based on \citet{intelligence2025pi}. Instead of learning the reward model from preferences, we learn from success/failure rewards where the reward at each timestep $t$ is:
\begin{equation}
    r_t = \begin{cases}
  0 & \text{if } t = T \text{ and trajectory succeeded} \\
  R_{\text{fail}} & \text{if } t = T \text{ and trajectory failed} \\
  -1 & \text{otherwise}
  \end{cases}
\end{equation}
where $R_{\text{fail}} = -100$.

The value target is the Bellman backup with no discount; we normalize them so the magnitude does not hurt the fit. 
\begin{equation}
    V(s_t) = \sum_{k \geq t} r_k
\end{equation}

We then condition the policy on the advantage
\begin{equation}
A_t = V(s_{t+1}) - V(s_t)
\end{equation}

Which is normalized in the same way as \Acronym when passed as input to the policy, for easily finding the correct values to condition at test time. The policy learning part follows the same recipe as \Acronym.

\textbf{Weighted Regression} We train the reward model on multi-dimensional preferences using the same recipe as for \Acronym. The difference is that instead of conditioning the policy on the reward scores, we use a different policy extraction method based on Advantage Weighted Regression \cite{peng2019advantage}. We average all the scores obtained from the reward model over the $L$ selected axes, and we modify the loss by weighting it by the average of the scores:

\begin{equation}
\label{eq:multi-axis-awr}
\mathcal{L}_{\text{AWR}}(\theta)
=
-\frac{\mathbb{E}_{\tau_i \sim \mathcal{D}}
\left[
\sum_{t=1}^{T_i}
\log \pi_\theta\!(
\mathbf{a}_i^t
\,\middle|\,
s_i^t) ~exp\left(\sum_k^{K_\pi} r_\phi(\tau_i \mid l_k)
\right)
\right]}{\sum_{j=0}^{\mid \mathcal{D}\mid}\sum_k^{K_\pi} r_\phi(\tau_j \mid l_k)},
\end{equation}

We refer to Table \ref{tab:task-results-appdx} and Table \ref{tab:sim-results-appdx} for all the detailed numbers on the real-world and simulation benchmarks. 

\section{Compute details}

The compute resources used for this project were one H100 for full finetuning of the policy and finetuning of the reward model. For the simulation experiments, smaller GPUs were used, generally NVIDIA RTX 4500.

\section{Detailed Results}

\subsection{Simulation Tasks}

Table \ref{tab:sim-results-appdx} reports the performance of each one of the baselines in simulation.

\begin{table}[!ht]
\centering
\caption{\footnotesize Simulation results comparing \Acronym against baselines across simulation environments. Multi-dimensional preferences as leveraged by \Acronym provide the best supervision.}
\label{tab:sim-results-appdx}
\small
\setlength{\tabcolsep}{6pt}
\begin{tabular}{lccc}
\hline
 & Object rearrangement & Bimodal square & Bimodal square \\ 
Method & (success) & (throughput) & inverted (throughput) \\
\hline
BC \cite{levine2016end}      & $0.04 \pm 0.02$ & $0.71 \pm 0.07$ & $0.25 \pm 0.14$ \\ 
Filtered BC \cite{emmons2021rvs} & $0.09 \pm 0.05$ & $0.71 \pm 0.02$ & $0.00 \pm 0.00$ \\

Weighted Regression \cite{peng2019advantage}  & $0.18 \pm 0.08$ & $0.83 \pm 0.08$ & $0.09 \pm 0.06$ \\ 
Advantage Conditioning \cite{intelligence2025pi}  & $0.17 \pm 0.06$  & $0.75 \pm 0.12$ & $0.08 \pm 0.04$ \\
Single Preferences    \cite{christianoHumanPref}    & $0.73 \pm 0.04$ & $0.67 \pm 0.08$ & $0.78 \pm 0.01$ \\ 
Single Preferences (matching labels)    & $0.79 \pm 0.06$ & $0.73 \pm 0.05$ & $0.74 \pm 0.05$ \\
\Acronym (Ours)           & $\mathbf{0.84 \pm 0.09}$ & $\mathbf{1.19 \pm 0.01}$ & $\mathbf{1.24 \pm 0.01}$ \\
\hline
\end{tabular}
\vspace{-1.5em}
\end{table}

Tables~\ref{tab:bimodal-square-detailed} and \ref{tab:bimodal-square-inverted-detailed} report the detailed per-metric breakdown for the Bimodal Square environment in the standard and inverted conditioning directions, respectively, including success rate, time to first step (lower is better), and throughput.

\begin{table}[H]
\centering
\caption{\footnotesize Detailed results on the Bimodal Square environment (standard direction). We report mean $\pm$ standard error over 3 seeds for success rate, time to first step (T1S, lower is better), and throughput.}
\label{tab:bimodal-square-detailed}
\small
\setlength{\tabcolsep}{6pt}
\begin{tabular}{lccc}
\hline
Method & Success & T1S & Throughput \\
\hline
Demo only        & $0.73 \pm 0.15$ & $347.71 \pm 7.14$  & $0.63 \pm 0.13$ \\
Filtered BC      & $0.92 \pm 0.03$ & $405.68 \pm 3.36$  & $0.68 \pm 0.02$ \\
Weighted Reg.    & $0.90 \pm 0.07$ & $333.77 \pm 20.45$ & $0.81 \pm 0.10$ \\
Single Pref      & $0.89 \pm 0.11$ & $395.93 \pm 1.80$  & $0.67 \pm 0.08$ \\
Single Pref. Matching Labels      & $0.95 \pm 0.05$ & $391.80 \pm 6.10$  & $0.73 \pm 0.05$ \\
Advantage Cond.      & $0.89 \pm 0.06$ & $356.31 \pm 31.81$ & $0.75 \pm 0.12$ \\
\Acronym~ (Ours) & $\mathbf{1.00 \pm 0.00}$ & $\mathbf{253.13 \pm 0.01}$ & $\mathbf{1.19 \pm 0.00}$ \\
\hline
\end{tabular}
\end{table}

\begin{table}[H]
\centering
\caption{\footnotesize Detailed results on the Bimodal Square environment (inverted direction). We report mean $\pm$ standard error over 3 seeds for success rate, time to first step (T1S, lower is better), and throughput.}
\label{tab:bimodal-square-inverted-detailed}
\small
\setlength{\tabcolsep}{6pt}
\begin{tabular}{lccc}
\hline
Method & Success & T1S & Throughput \\
\hline
Demo only        & $0.27 \pm 0.15$ & $330.12 \pm 12.37$ & $0.25 \pm 0.14$ \\
Filtered BC      & $0.00 \pm 0.00$ & $0.00 \pm 0.00$    & $0.00 \pm 0.00$ \\
Weighted Reg.    & $0.09 \pm 0.08$ & $292.26 \pm 37.31$ & $0.09 \pm 0.06$ \\
Advantage Cond.      & $0.09 \pm 0.06$ & $317.99 \pm 60.07$ & $0.08 \pm 0.04$ \\
Single Pref. Matching Labels       & $0.96 \pm 0.02$ & $390.67 \pm 18.22$ & $0.74 \pm 0.05$ \\
Single Pref      & $0.97 \pm 0.01$ & $376.50 \pm 4.96$  & $0.78 \pm 0.01$ \\
\Acronym~ (Ours) & $\mathbf{1.00 \pm 0.00}$ & $\mathbf{242.23 \pm 0.06}$ & $\mathbf{1.24 \pm 0.00}$ \\
\hline
\end{tabular}
\end{table}

\subsection{Real World Tasks}
Table~\ref{tab:task-results-appdx} reports per-task success rates for real-world experiments across all methods.

\begin{table}[H]
\centering
\caption{\footnotesize Success rate per task and method. We report the mean $\pm$ standard error.}
\label{tab:task-results-appdx}
\resizebox{\linewidth}{!}{%
\begin{tabular}{lccccc}
\hline
Method & Setup Table & Plate Toast & Fold Shorts & Place Cube into Bowl & Average\\
\hline
BC  & $0.68 \pm 0.05$ & $0.15 \pm 0.08$ & $0.10 \pm 0.07$ & $0.33 \pm 0.10$ & $0.31 \pm 0.04$ \\
Filtered BC        & $0.65 \pm 0.04$ & $0.30 \pm 0.10$ & $0.25 \pm 0.10$ & $0.29 \pm 0.10$ & $0.37 \pm 0.04$\\
Single Preference (match pairs) & $0.64 \pm 0.06$ & $0.00 \pm 0.00$ & $0.25 \pm 0.10$ & $0.43 \pm 0.11$ & $0.33 \pm 0.04$\\
Single Preference (match comparisons)  & $0.77 \pm 0.03$ & $0.10 \pm 0.07$ & $0.30 \pm 0.10$ & $0.19 \pm 0.09$ & $0.34 \pm 0.04$\\
\Acronym~ (Ours)   & $\mathbf{0.94 \pm 0.02}$ & $\mathbf{0.70 \pm 0.10}$ & $\mathbf{0.55 \pm 0.11}$ & $\mathbf{0.81 \pm 0.09}$ & $\mathbf{0.75 \pm 0.04}$\\
\hline
\end{tabular}}
\end{table}

\subsubsection{Steerability of the policies}
In Figure \ref{fig:steerability}, we depict the results from Table \ref{tab:task-results-appdx} for the place cube into bowl task. More concretely, we show an overlay of the commanded bowl and the bowl where it was placed. We observe that \Acronym can achieve steerability by changing at test-time which reward dimension we want the policy to optimize, while the baselines do not.

\begin{figure}[h!]
    \centering
    \includegraphics[width=\linewidth]{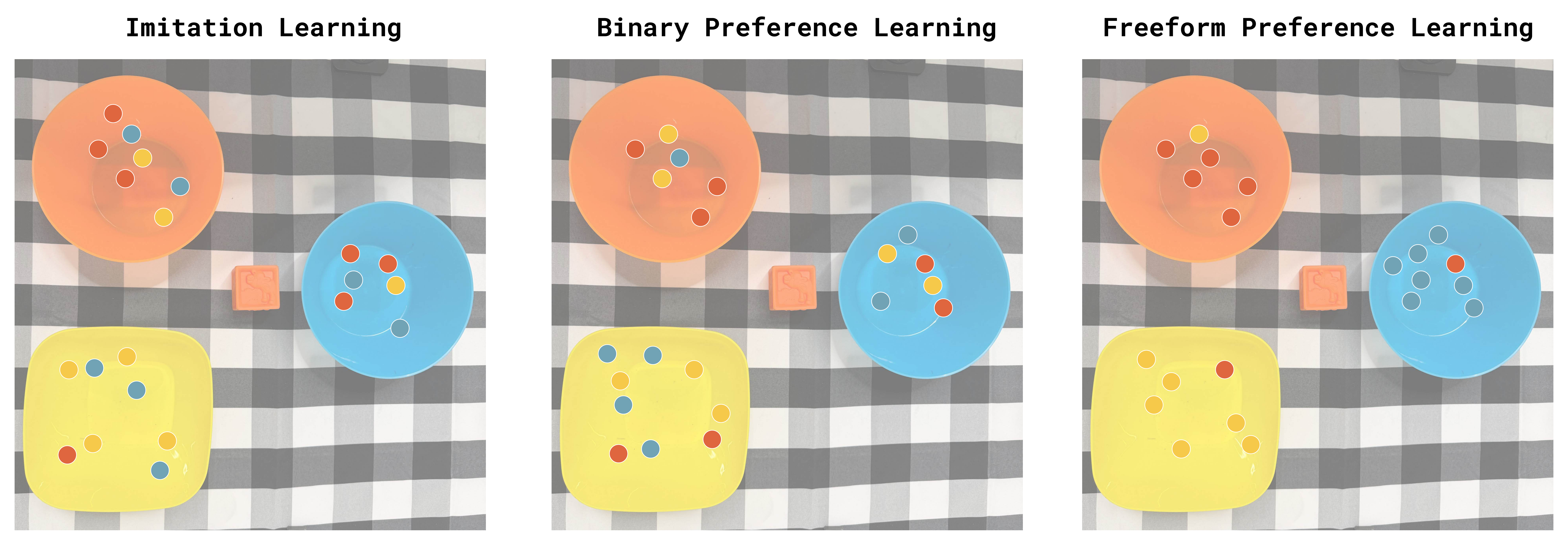}
    \caption{ \footnotesize Steerability Test. Each dot represents the block's final position, with the color corresponding to the target bowl's color. \Acronym achieves steerability by placing the cube in the commanded bowl most of the time, whereas the placement is close to random for the other two baselines. Steerability is achieved for \Acronym by prompting, at test time, the reward it needs to optimize.}
    \label{fig:steerability-appdx}
\end{figure}

\subsubsection{Iterative Improvement}

\begin{figure}[H]
    \centering
    \begin{subfigure}[t]{0.48\linewidth}
        \centering
        \includegraphics[width=\linewidth]{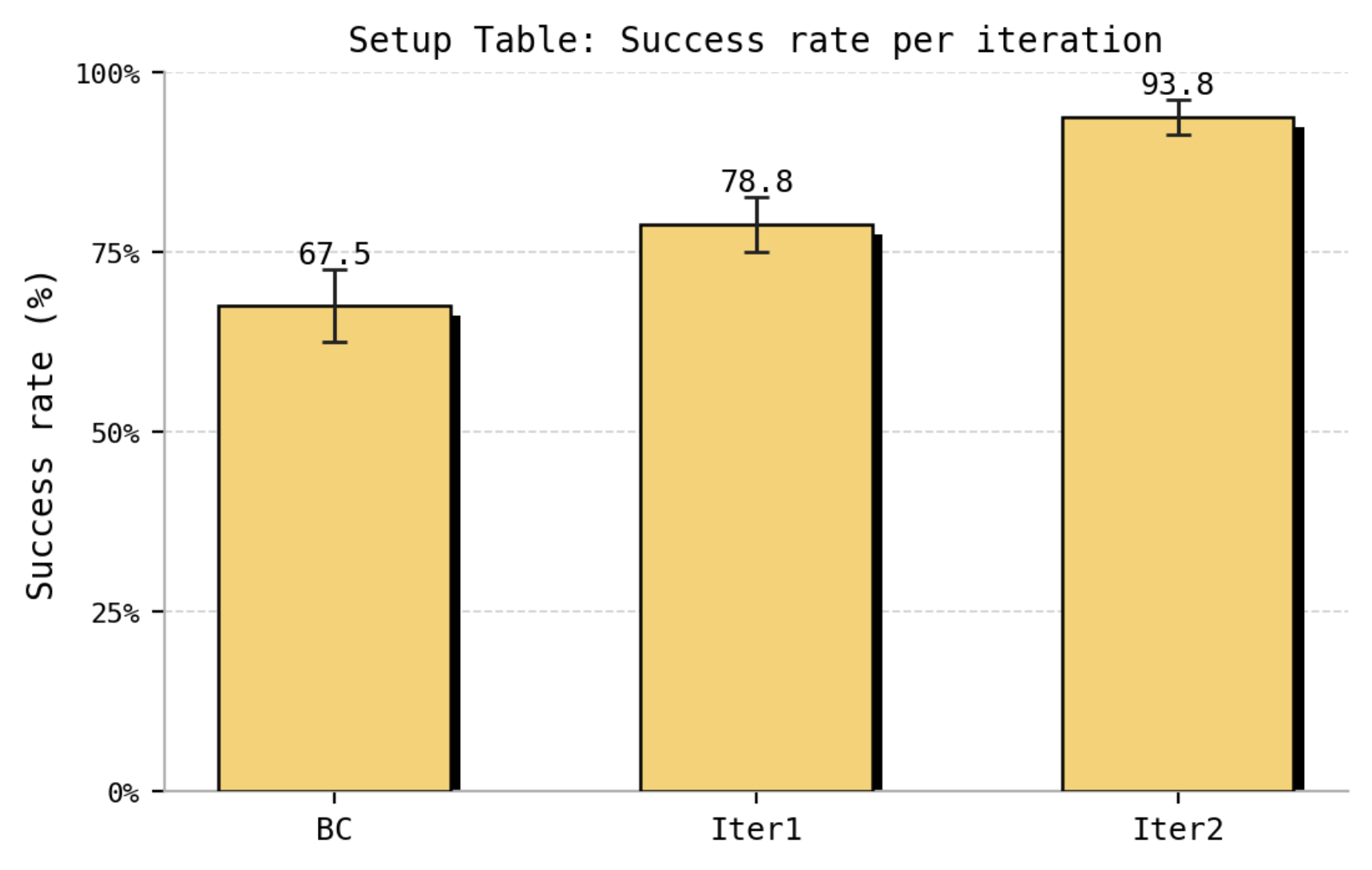}
        \caption{\footnotesize Policy success rate per iteration on the \emph{setup table} task.}
        \label{fig:setup-success}
    \end{subfigure}
    \hfill
    \begin{subfigure}[t]{0.48\linewidth}
        \centering
        \includegraphics[width=\linewidth]{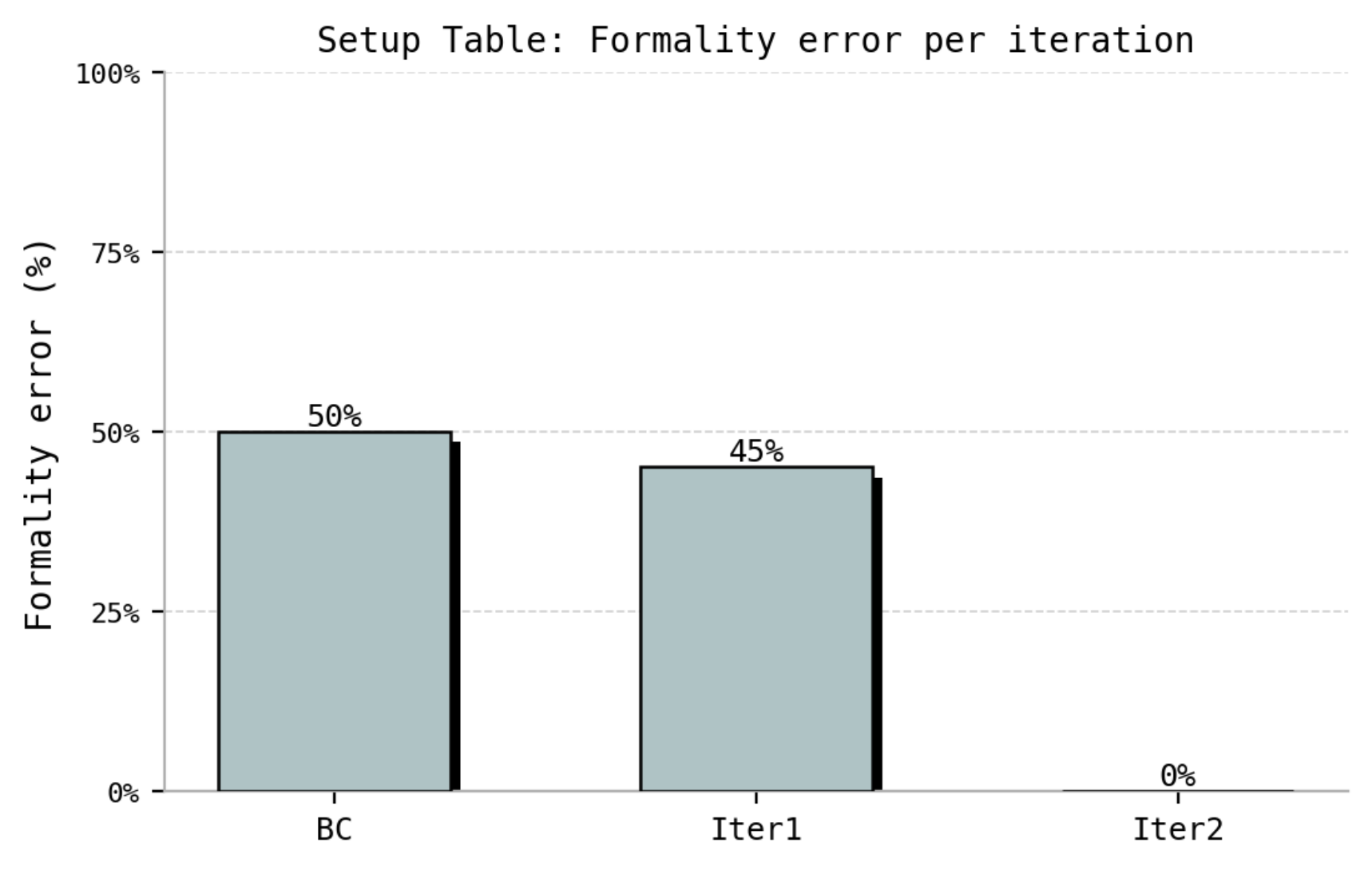}
        \caption{\footnotesize Formality error rate.}
        \label{fig:error-rate}
    \end{subfigure}
    \caption{\footnotesize Iterative analysis on the \emph{setup table} task: (\subref{fig:setup-success}) policy success climbs across iterations, and (\subref{fig:error-rate}) the formality error rate falls to 0.}
    \label{fig:setup-iterative-analysis}
\end{figure}

Figure~\ref{fig:setup-iterative-analysis} shows how \Acronym~ improves policy between iterations of the Setup Table task, compared to BC. We report the task success rate (higher is better) and the formality error rate (lower is better).

\begin{figure}[H]
    \centering
    \includegraphics[width=0.6\linewidth]{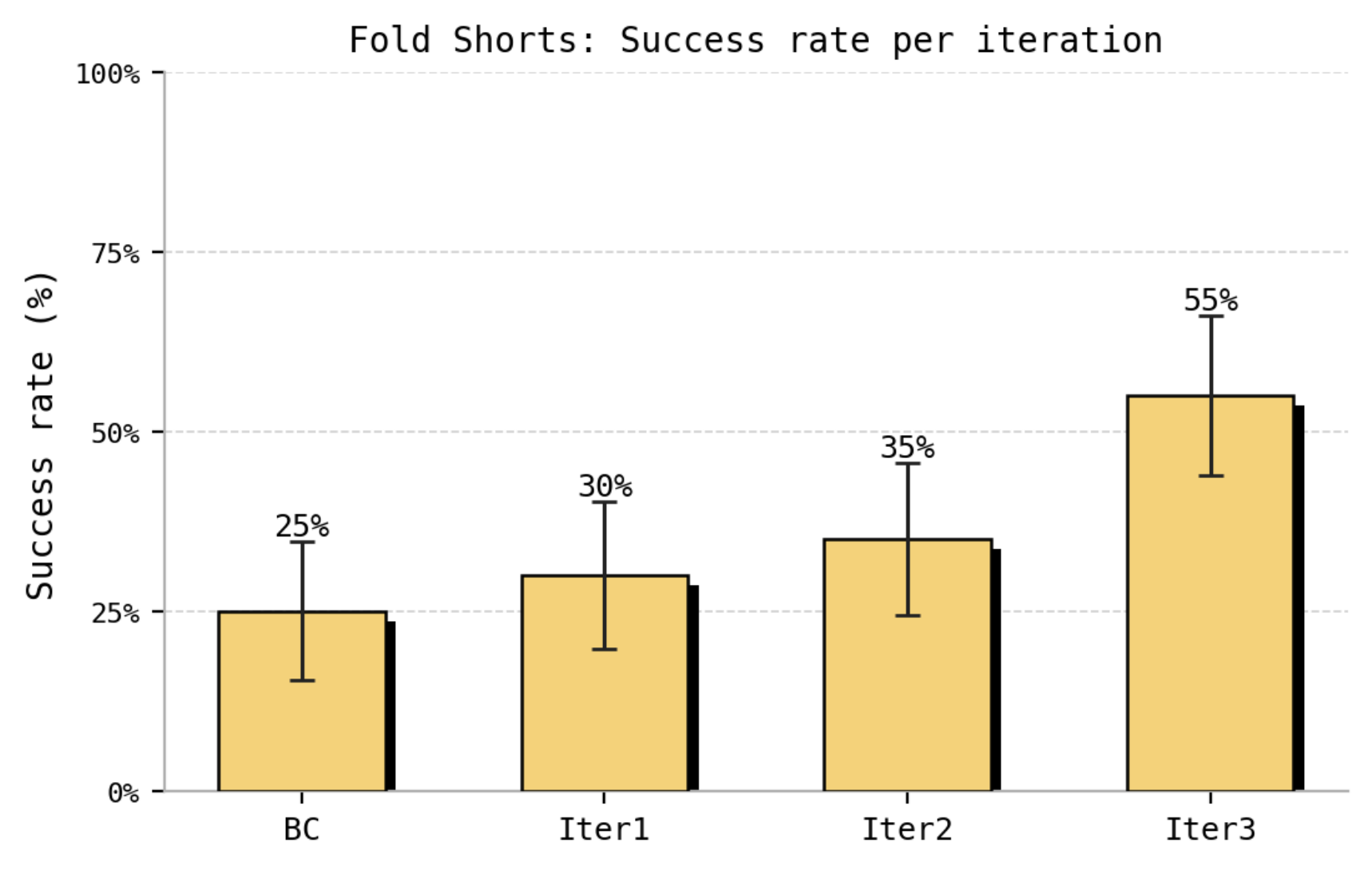}
    \caption{\footnotesize Policy success rate per iteration on the \emph{fold shorts} task.}
    \label{fig:fold-iterative-analysis}
\end{figure}

Figure~\ref{fig:fold-iterative-analysis} shows the analogous progression on the iterative Fold Shorts task, where preferences are collected in \emph{freeform} format. Policy success increases across iterations relative to the BC baseline.

\begin{landscape}
\begin{figure}
    \centering
    \includegraphics[width=\linewidth,height=0.85\textheight,keepaspectratio]{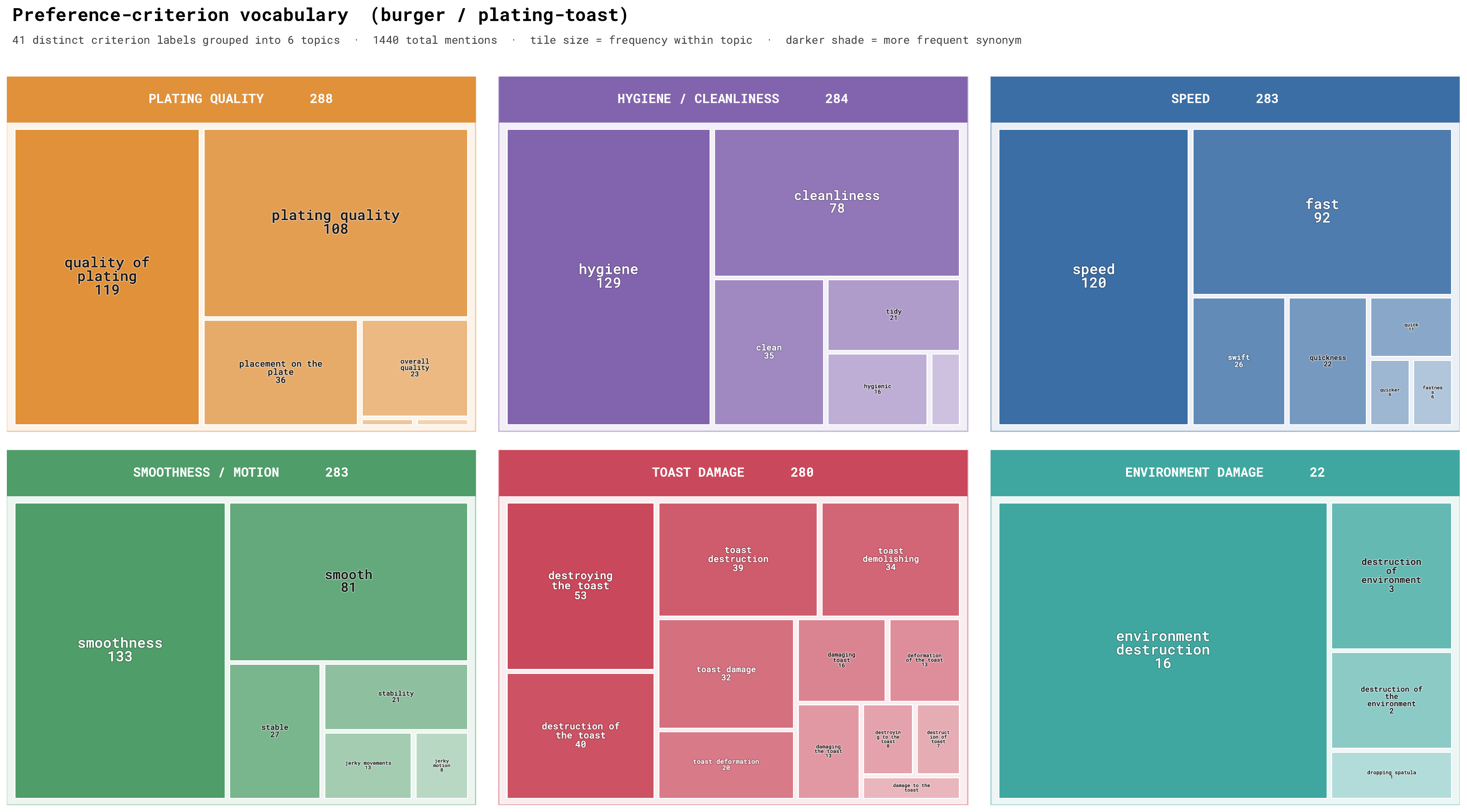}
    \caption{\footnotesize Freeform preference labels collected for the \emph{plate toast} task grouped by topic. Tile size and shade reflect label frequency. 
    }
    \label{fig:prefdetails}
\end{figure}
\end{landscape}

\end{document}